%% file: main.tex
\documentclass[preprint,9pt]{elsarticle}
\usepackage{geometry}
\usepackage{color}

\usepackage{dblfloatfix}  
\usepackage{booktabs}
\usepackage{amsmath, bm}
\usepackage{amsthm}
\theoremstyle{definition}
\newtheorem{definition}{Definition}[section]
\usepackage{graphicx}
\usepackage{longtable}
\usepackage{multirow}

\usepackage[table]{xcolor}
\usepackage[flushleft]{threeparttable}
\usepackage{tikz}
\usepackage{soul}
\usepackage{tikzpeople}
\usepackage{algorithm} 
\usepackage{algpseudocode} 
\usepackage{amsmath}
\DeclareMathOperator*{\argmax}{arg\,max}

\usetikzlibrary{matrix,shapes.geometric, arrows.meta,positioning,fit, spy, intersections, arrows,shapes.multipart, calc}
\tikzstyle{line}=[draw]
 
\journal{Applied Soft Computing}
\usepackage{graphicx}
\usepackage{subfig}
\usepackage{tikz,tikz-3dplot}
\usetikzlibrary{3d,decorations.text,shapes.arrows,positioning,fit,backgrounds, shapes, arrows,arrows.meta, intersections}
\usetikzlibrary{decorations.markings, decorations.text}
\usetikzlibrary{shapes.geometric,positioning}
\usepackage{pgfplots}
\usepackage[alpine]{ifsym}
\usepgfplotslibrary{fillbetween}
\usetikzlibrary{calc}
\usepackage[export]{adjustbox}
\usepackage{gnuplottex}
\usepackage{tikz-3dplot}
\usepackage{lscape}
\usepackage[normalem]{ulem}
\definecolor{cerisepink}{rgb}{0.93, 0.23, 0.51}
\definecolor{color1}{RGB}{255, 253, 204}

\begin{document}
\begin{frontmatter}

\title{Fuzzy Temporal Convolutional Neural Networks in P300-based Brain-Computer Interface for Smart Home Interaction}


\author[inst1,inst3]{Christian Flores Vega}
\affiliation[inst1]{organization={Universidad de Ingeniería y Tecnología - UTEC},
            city={Lima},
            country={Peru}}
\affiliation[inst3]{organization={University of Campinas},
            city={São Paulo},
            country={Brazil}}

\author[inst1]{Jonathan Quevedo}
\author[inst1]{Elmer Escandón}

\affiliation[inst2]{organization={University of Essex},
            city={Colchester},
            country={United Kingdom}}

\author[inst2]{Mehrin Kiani}

\affiliation[inst4]{organization={Nantong University},
city={Jiangsu},
country={China}}

\author[inst4]{Weiping Ding}

\author[inst2,inst5]{\\Javier Andreu-Perez$^*$}
\ead{javier.andreu@essex.ac.uk, jandreu@ujaen.es}

\affiliation[inst5]{organization={University of Jaen}, 
            city={Jaén},
            country={Spain}}

\begin{abstract}
The processing and classification of electroencephalographic signals (EEG)  are increasingly performed using deep learning frameworks, such as convolutional neural networks (CNNs), to generate abstract features from brain data, automatically paving the way for remarkable classification prowess. However, EEG patterns exhibit high variability across time and uncertainty due to noise. It is a significant problem to be addressed in P300-based Brain Computer Interface (BCI) for smart home interaction. It operates in a non-optimal natural environment where added noise is often present and is also white. In this work, we propose a sequential unification of temporal convolutional networks (TCNs) modified to EEG signals, LSTM cells, with a fuzzy neural block (FNB), we called EEG-TCFNet. Fuzzy components may enable a higher tolerance to noisy conditions. We applied three different architectures comparing the effect of using block FNB to classify a P300 wave to build a BCI for smart home interaction with healthy and post-stroke individuals. Our results reported a maximum classification accuracy of $98.6 \%$ and  $74.3\%$ using the proposed method of EEG-TCFNet in subject-dependent strategy and subject-independent strategy, respectively. Overall, FNB usage in all three CNN topologies outperformed those without FNB.
In addition, we compared the addition of FNB to other state-of-the-art methods and obtained higher classification accuracies on account of the integration with FNB. 
The remarkable performance of the proposed model, EEG-TCFNet, and the general integration of fuzzy units to other classifiers would pave the way for enhanced P300-based BCIs for smart home interaction within natural settings.
\end{abstract}



\end{frontmatter}
\section{Introduction} \label{sec:Introduction}
Brain-computer interface (BCI) enables controlling a device (computer) using commands decoded from brain signals acquired from a given neuroimaging technology \cite{intro_1}. BCIs provide an alternative mechanism for communicating a user's intent, recognized from their brain signals, to control a given device. The interest in BCI development arises from the profound implications it (BCI) can have to assist people with debilitating muscle disabilities by giving them an alternative communication option independent of voluntary muscle movement \cite{sun2019advanced, tang2020motor}. 

The neuroimaging technology most often used in BCIs, for recording brain signals is Electroencephalography (EEG)  \cite{vavreka2020evaluation}. EEG is non-invasive, portable, and allows a participant to record his brain activity in more natural settings, such as sitting upright, while performing a motor or cognitive task. In addition, the widespread EEG usage in BCI applications is because of EEG's high temporal resolution (in the order of milliseconds), which gives an almost real-time capture of the brain activity, making it ideal for BCI applications \cite{Lotte_2018_EEG_BCIreview}. Despite having a low spatial resolution [6], the suitability of EEG is not affected for BCI applications. The brain responses most commonly used for EEG based BCI applications, such as event-related potentials (ERP), do not need to be mapped to specific anatomical locations to infer brain activity.. 

An ERP is a time-locked brain response that is elicited in response to a specific motor or cognitive task \cite{Sur_2009_ERP}. ERPs are typically defined with respect to latency (time post-stimulus presentation) and the amplitude of the potential such as the P300. The P300 is a positive peak observed after 300 milliseconds of stimulus presentation. The P300 wave can be elicited using an \emph{oddball} paradigm where the user is requested to respond to the infrequent (oddball) target stimulus. A reduced amplitude of P300 can indicate broad neurobiological vulnerabilities such as alcohol dependence \cite{Patrick_2006_P300}.

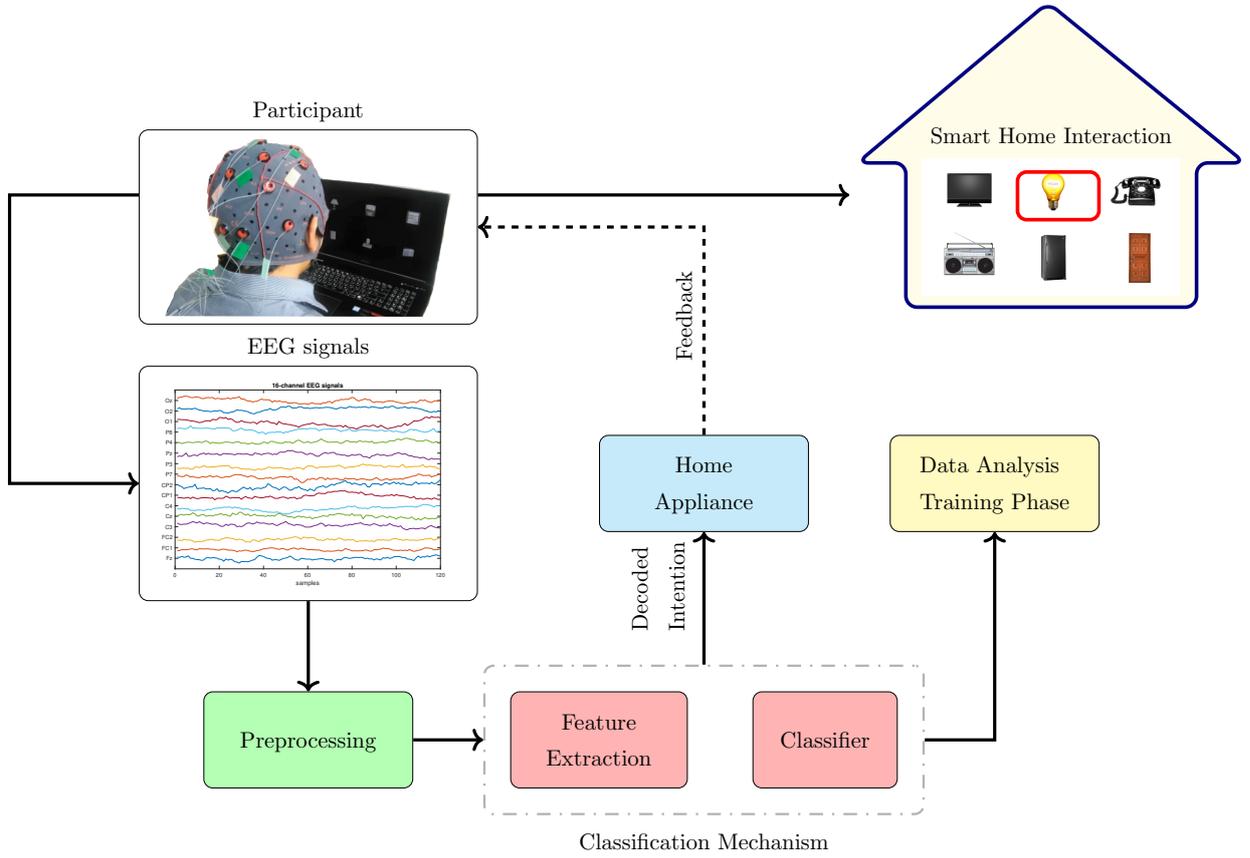
\begin{figure}[!t]
\centering
\scalebox{0.85}{\input{Figures/BCI_schematic}}
\caption{A schematic of a closed-loop EEG-based Brain-computer interface (BCI) paradigm. In this illustration, the EEG-based BCI is undertaking the P300 smart home interaction, paradigm for selecting which commonly found home appliances the user wants to interact with. The brain activity is recorded by the electrodes on the EEG cap placed on a participant's head. In order to improve the signal-to-noise ratio (SNR), the EEG signals are first pre-processed (,such as signal normalization and enhancement by removing direct current \cite{Zhang_2020_EEGreview_Deeplearning}). The pre-processed EEG signals are then used for feature extraction typically from the time domain (statistical features such as mean, variance), or frequency domain (such as Fourier transform), and more recently from the time-frequency domain (such as discrete wavelet transform). The classifier learns from the training data based on the extracted features to discriminate between the user's intentions such as which device they are focusing on. The decoded intention of the user is then displayed back at the screen as a form of feedback to the user as to whether the correct intention has been decoded successfully.}
\label{fig:EEG_BCI_flowchart}
\end{figure}

The P300 based BCI for smart home interaction allows people to interact (such as switch on or off) with commonly found devices at home using their brain signals. The P300 is evoked in the participants by causing a surprise event using an oddball paradigm \cite{cortez2021smart}.  A general schematic for a P300 based BCI for smart home interaction is shown in Fig. \ref{fig:EEG_BCI_flowchart}. A typical display of the devices commonly found in homes is shown in Fig. \ref{fig:EEG_BCI_flowchart}. The EEG electrodes record the P300 as the participants focus on a particular device they want to switch on/off. 

A key component in any BCI paradigm is the classification model, also shown in Fig. \ref{fig:EEG_BCI_flowchart}, that decodes the intent of participants based on the information in the brain signals. Over the past decade, the marked improvement in BCI paradigms can be attributed to the computational prowess of the classification models to successfully decode the EEG signals \cite{Cecotti_2011}. In this regard, the most noteworthy improvement in classification models is the evolution of neural networks into convolutional neural networks (CNNs). Owing to their hierarchical structure that can learn abstract features from the input data, CNNs have consistently given remarkable classification accuracy over a range of complex real-life classification problems such as image classification \cite{LeCun_2015_DeepLearning}. Motivated from the success of CNNs in classification problems, we propose the use of Temporal Convolutional Networks (TCN) in tandem with long-short term memory (LSTM) networks, which can memorise signal variation and signal characterisation, and fuzzy neural block (FNB) as the last layer before classification.

We tested our proposed deep learning model's efficacy with FNB to classify the target action based on the P300 smart home interaction using EEG signals from our smart home control experiment  \cite{cortez2021smart, P300_Db}. The implementation reaches the highest accuracy of $98.58\% \pm 0.49\%$ to subject-dependent classification with leave-one-session-out cross-validation and $74.21\% \pm 1.72\%$ to subject-independent classification with leave-one-subject-out-cross-validation.

The main contributions of the present work are: 

\begin{itemize}
\item A comprehensive study of temporal CNN decoding approaches for a BCI-P300 paradigm of smart home interaction in a natural environment with healthy and stroke individuals using deep learning.

\item A proposal of a CNN architecture, EEG-TCFNet, for P300 wave-based smart home interaction in natural (non-shielded) environments. The results indicate higher accuracy of the proposed model, EEG-TCFNet, versus state-of-the-art methods with similar topologies.  


\item An assessment of the inclusion of Fuzzy Neural Block (FNB) with different deep learning topologies.  The use of FNB in the deep learning architecture was motivated by the fact that the P300 BCI application of this experiment was performed in real settings with natural surroundings, not in a laboratory with control settings. Therefore, there could be more uncertainty in the recordings in these scenarios as more frequent potential noisier factors and distractors are present. Consequently, we hypothesized that accounting for the fuzziness in the model pattern parameters could help improve the ecological validity of P300-based BCI for smart home settings. The results reported that the addition of the FNB improves the BCI performance in the smart home interaction scenario of this study.

\end{itemize}

The rest of the paper is organised as follows. Section 2 outlines the related works, with section 3 presenting a background of the classification components used in the present work. In section 4, first, the experimental paradigm and the EEG signals' preprocessing is presented. A comparison of the state-of-the-art methods for the classification of the signals is presented next, followed by a detailed description of the proposed network. We present the results in section 5, and finally, section 6 delineates the conclusions and future work.
\section{Related Work}
This section outlines the related works of the different components of our proposed model, EEG-TCFNet. For mobility impaired or lock-in individuals, interacting with the environment is the first step towards some level of independence. The P300-based visual stimulation is an effective paradigm for BCI utilised as speech or word spellers. 
In our previous work \cite{cortez2021smart}, we have noted the paradigm worked well for healthy and post-stroke individuals with an approximate 90\% accuracy for P300-based BCI for smart home interaction.
Testing more complex classifiers such as deep neural networks \cite{cortez2020under}, we observed that the accuracy of a P300-based speller increased by 7\% on the worse subject, and the maximum accuracy went slightly up the 90\% previously achieved. A similar boosting in performance has been reported in other P300-based works when introducing deep learning approaches. In \cite{cecotti20193d} 2D and 3D based CNN were assessed in their performance for P300 event-related potential detection. Among the four 3D models and two 2D models, the higher dimension model reached a \(92.8\% \pm 5\%\) AUC efficiency. Recently some other works have tested the recognition capabilities of recurrent deep architectures that may also exploit time-course patterns. For instance, in \cite{ditthapron2019universal} is presented an architecture based on CNN + LSTM (ERPENet) that achieves a consistent accuracy between 79\% and 88\% and outperforms other non-recurrent methods. 

From this body of evidence, we hypothesise that the inclusion of novel layers in a CNN could improve our application's performance. A new temporal CNN is denoted in the literature as Temporal CNN (TCNN) \cite{bai2018empirical} has outperformed CNN with classical recurrent layers in pattern recognition in signal processing  \cite{lin2020temporal, ingolfsson2020eeg}. Likewise, a recent trend is the inclusion into CNN architectures of fuzzy logic components \cite{rovzman2020privacy, sharma2019fuzzy}. A drawback of learning EEG patterns with deep learning is the assumption that these patterns are noise-free and not affected by non-stationarity, therefore ignoring the uncertainty. In this regard, fuzzy sets and systems have shown performance gains in noisy BCI problems \cite{andreu2016}.

As discussed, although much effort has been dedicated to classifying the target action based on the P300 BCI, some points that need to improve:
\begin{enumerate}[(i)]
\item Enhance the operation of BCI (accuracy and bitrate) based on the P300 paradigm in non-optimal environments due to a BCI should operate in real-life conditions.
\item Improving model decoding capabilities to provide a balanced performance across subject-specific, subject-independent, healthy and patient settings. 
\end{enumerate}

In this work, we expand the aforementioned state-of-the-art deep learning methods in BCI with a FNB and evaluate them in our target BCI interaction speller for smart homes. By the time of this research, no other work has combined DNN with fuzzy systems in P300-based BCI systems, and little is known about the application of this paradigm outside of lab conditions such as home environments.

\section{Background} \label{sec:BG}
This section summarises state-of-the-art classification models of LeNet  \cite{lecun1998gradient}and EEG-TCNet \cite{ingolfsson2020eeg} which have previously been used in P300-based BCI systems. In addition, we also delineate the LSTM network \cite{Graves_2005_LSTM}. In the present work, we compare the aforementioned networks' performance with and without the proposed FNB. 




\subsection{LeNet Network}
The LeNet network \cite{lecun1998gradient} is one of the simplest CNN used for the recognition of handwritten digits. The LeNet consists of 6 layers that involve two 2D convolutional layers, a 5x5 kernel size, a rectified linear (ReLU) activation function per layer, and a softmax activation function output layer. 
The architecture of the LeNet is as noted in Table \ref{tab:lenet_two_networks}.

\begin{table}[!b]
\centering
\caption{The LeNet network structure consists of two 2 dimensional (2D) convolutional (Conv) layers, average pooling and a rectified linear (ReLU) activation function per layer.}
\begin{tabular}{clccc}
\hline
\textbf{Layer} & \textbf{Type} & \textbf{\# Filters} & \textbf{Kernels} & \textbf{Output} \\ \hline
\multirow{3}{*} {$L1$} & Input & - & - & (16,120,1) \\
 & Conv2D & 6 & (5,5) & (12,116,6) \\
 & AveragePool2D & - & (2,2) & (6,58,6) \\ \hline
\multirow{2}{*} {$L2$} & Conv2D & 16 & (5,5) & (2,54,16) \\
 & AveragePool2D & - & (2,2) & (1,27,16) \\ \hline
$L3$ & Flatten & - & - & 432 \\ \hline
$L4$ & Dense & - & - & 120 \\ \hline
$L5$ & Dense & - & - & 84 \\ \hline
$L6$ & Dense & - & - & 2 \\ \bottomrule
\end{tabular}
\label{tab:lenet_two_networks}
\end{table}
\subsection{EEG-TCNet}
The EEG-TCNet \cite{ingolfsson2020eeg} consists of the EEG-Net \cite{lawhern2018eegnet} combined with a Temporal Convolution Network (TCN). The structure begins with a temporal convolution ($L1$) to serve as frequency filters. Then, it uses a depth-wise convolution ($L2$) to learn spatial filtering from each frequency band. Lastly, a separable convolution ($L3$) summarizes the feature map individually, resulting in a combination of the maps afterward.  A zero-padding scheme is applied to preserve the EEG's temporal characteristics and learn spatial filtering properties,  maintaining the channel dimensions (16).

In addition, a batch normalization with an exponential linear unit (ELU) activation function is applied since it provides greater accuracy than other functions \cite{ingolfsson2020eeg}. The ELU can be mathematically defined as shown in  eq. (\ref{eq.ELU})  \cite{Wang_2020}:   
\begin{align}
f(x)=\begin{cases}
                   x, & x\geq 0  \\
    \alpha (e^{x}-1), & x < 0  \\
\end{cases}
\label{eq.ELU}
\end{align}
where x is the input eigenvalue, and $\alpha$ is the hyperparameter that controls the saturation of ELU.

A dropout probability with the value of 0.5 in the TCN is included to prevent overfitting. After the fourth layer ($L4$), the extracted features are input data for the TCN due to existing temporal characteristics. Therefore, additional TCNs can extract further temporal information. The TCN includes six 2x2 filters with a classification layer using a softmax activation function.

\begin{table}[!b]
\centering
\caption{The architecture of the EEG-TCNet \cite{ingolfsson2020eeg} consists of an EEG-Net and a temporal convolution network (TCN).}
\begin{tabular}{clccc}
\hline
\textbf{Layer} & \textbf{Type} & \textbf{\# Filter} & \textbf{Kernel} & \textbf{Output} \\ \hline
\multirow{3}{*}{$L1$} & Input & - & - & (16,120,1) \\ \cline{2-5} 
 & Conv2D & 8 & (1,20) & \multirow{2}{*}{(12,120,8)} \\
 & BatchNorm & - & - &  \\ \hline
\multirow{5}{*}{$L2$} & DepthwiseConv2D & 16 & (16,1) & \multirow{3}{*}{(1,120,16)} \\
 & BatchNorm & - & - &  \\
 & Activation (ELU) & - & - &  \\ \cline{2-5} 
 & AveragePool2D & - & (1,4) & \multirow{2}{*}{(1,30,16)} \\
 & Dropout & - & - &  \\ \hline
\multirow{5}{*}{$L3$} & SeparableConv2D & 8 & (1,6) & \multirow{3}{*}{(1,30,8)} \\
 & BatchNorm & - & - &  \\
 & Activation (ELU) & - & - &  \\ \cline{2-5} 
 & AveragePool2D & - & (1,5) & \multirow{2}{*}{(1,6,8)} \\
 & Dropout & - & - &  \\ \hline
$L4$ & Flatten & - & - & (1,48) \\ \hline
$L5$ & TCN & 6 & (2,2) & 6 \\ \hline{$L6$} & Dense & - & - & 2 \\ \bottomrule
\end{tabular}
\label{tab:eeg_tcnet_combined}
\end{table}
\subsection{LSTM based CNNs}
The LSTM layer\cite{Graves_2005_LSTM} is a type of recurrent neural network that can learn long-term dependencies within the input data. When the LSTM layer is stacked to a CNN architecture, it can effectively extract the temporal features of the brain signals. 

At the core of the LSTM is the cell state which can be modified by adding or removing information from the cell state. The addition or removal of the information from the cell states is regularised using structures called gates. The LSTM networks are based on stacked blocks consisting of three gates which are named the input gate, output gate, and forget gate.
The aforementioned control cells are described by the following equations:
\begin{align}
     i&=\sigma(W_{i}x_{t}+U_{i}h_{t-1}+b_{i})  \\
     f&=\sigma(W_{f}x_{t}+U_{f}h_{t-1}+b_{f})  \\
     o&=\sigma(W_{o}x_{t}+U_{o}h_{t-1}+b_{o}) \\
     \tilde{c}&=W_{c}x_{t}+U_{c}h_{t-1}+b_{c}\\
     c_{t}&=f {\odot} c_{t-1}+i {\odot} \tilde{c}\\
     h_{t}&=o {\odot} c_{t}\\
    \sigma(x)&=\frac{1}{1+e^{-x}} \
\end{align}
where $W$, $U$ and $b$ represent sets of learnable parameters to control each gate. $x$, $h$, $i$, $f$, $o$ and $c$ represents input, output, input gate, forget gate, output gate and memory cell state, respectively. $\odot$ represents element-wise product. $t$ represents the data as the time series.

\section{Methodology}
In the present work, we apply P300-based interfacing to interface with appliances in a smart home. The user will receive visual stimulation of a picture of a particular item, and the BCI will have to decode the likely P300 response elicited from the different items that flash. A schema illustrating the smart home P300-based BCI is shown in Fig. \ref{fig:EEG_BCI_flowchart}.

\subsection{Participants and EEG Data Acquisition}

\begin{figure}[t]
	\centering
\subfloat[The demographic information of the participants.]{
\begin{tabular}[b]{l l l l}
    \toprule
\textbf{Sub.} & \textbf{Age} & \textbf{Gender} & \textbf{Diagnosis}  \\
        \midrule
      S01  & 33 & Male & Healthy \\
      S02  & 21 & Male & Healthy \\
      S03  & 20 & Male & Healthy  \\
      S04  & 21 & Male & Healthy    \\
      S05  & 24 & Male & Healthy   \\
      S06  & 29 & Male & Healthy  \\
      S07  & 20 & Male & Hemorrhagic post-stroke    \\
      S08  & 52 & Female & Ischemic post-stroke   \\
      S09  & 55 & Male  & Ischemic post-stroke  \\
\bottomrule
\end{tabular}
}
\hspace{.1cm} 
\subfloat[Electrode positions according to the 10-10 system.]{\includegraphics[scale=0.28]{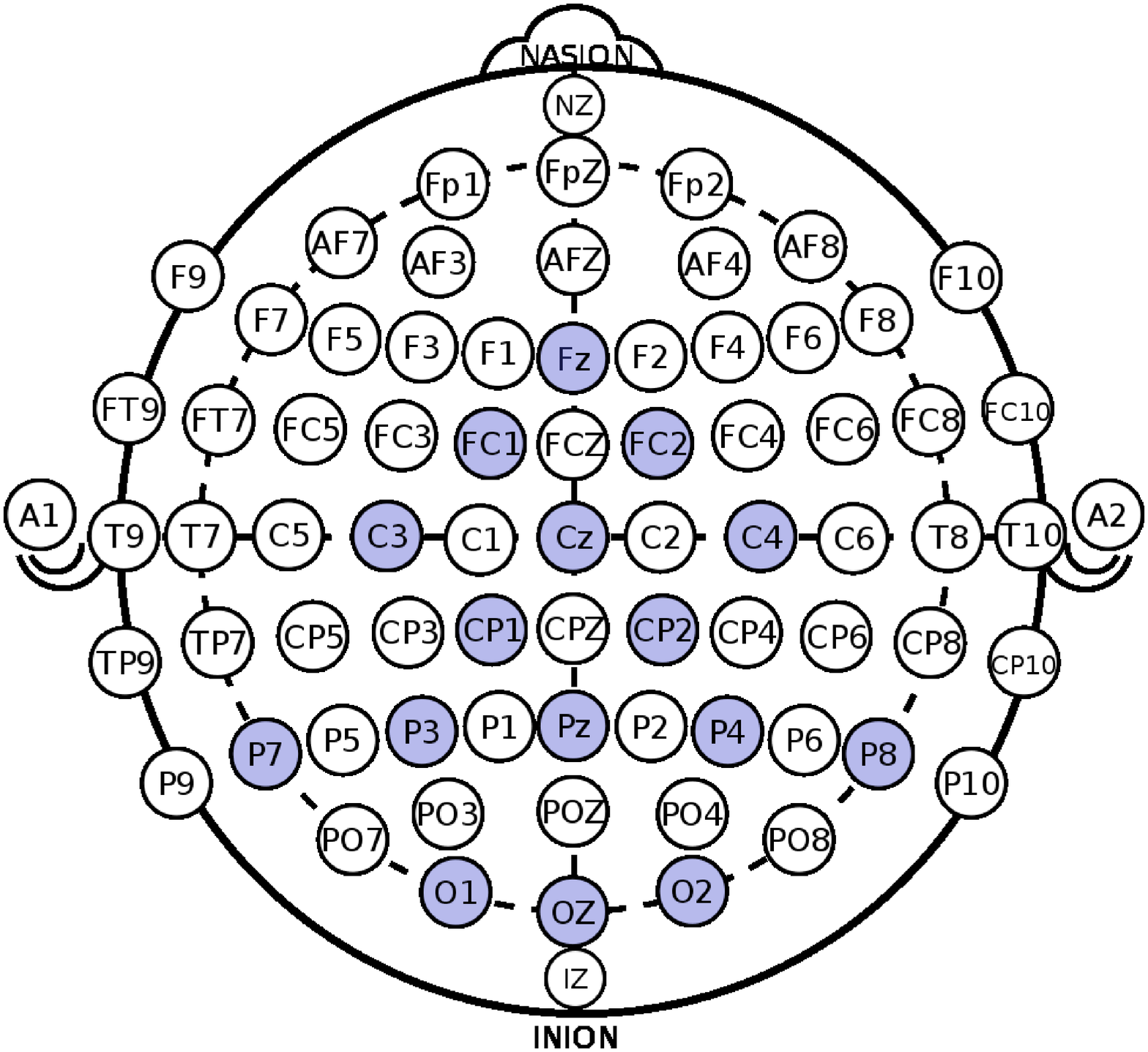}}

\subfloat[Stimulus presentation timeline.]{\includegraphics[width=1\columnwidth]{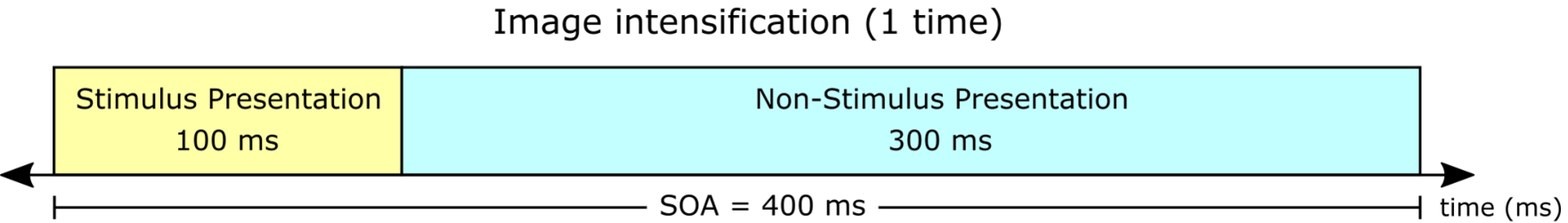}}
\caption{(a) A total of 9 participants performed in the P300 based smart home interaction BCI experiment. Of these 9 participants, six were healthy, and the remaining three had underlying health conditions. (b) The electrodes position as per the 10-10 system with the electrodes used in the experiment colored light blue. (c) A timeline depicting stimulus-onset asynchrony (SOA) of 400ms with the stimulus presentation of duration 100ms, followed by a non-stimulus presentation of duration 300 ms.}
\label{fig:10_20_system} 
\end{figure}

The \textit{Speller} test was performed on nine subjects, see further information in Fig. \ref{fig:10_20_system} (a), divided into two groups: control and evaluation. The control group includes only healthy subjects from S01 to S06, while subjects S07 to S09 are patients with a post-stroke condition. Subject S07 and S08 presented mild aphasia, especially S08 showed paresis in her upper limbs, and Subject S09 manifest severe apraxia \cite{sergio_P300_1}. All the volunteers agreed to participate in the experiment through a written consent reviewed and accepted by the Ethics Committee from the Universidad Peruana Cayetano Heredia. The document stated the study's academic objectives and the terms that ensure the anonymity of the participant.



For the EEG signal acquisition, a 16-channel g.USBamp amplifier (g.tec medical engineering  GmbH, Austria) system measures the signal values with a sampling rate of 2400Hz \cite{P300_Db}, and bipolar electrodes were positioned on the scalp surface area following the 10-10 system on the positions, see Figure \ref{fig:10_20_system} (b) : Fz, FC1, FC2, C3, Cz, C4, CP1, CP2, P7, P3, Pz, P4, P8, O1, O2, and Oz. Similarly, the ground and reference electrodes were placed on the right mastoid and left earlobe, respectively.  

\subsection{Experimental Home Interaction Environment}
In order to recreate home environments that emulate real-life conditions to end-users in natural settings, the experiment was performed in a non isolated or shielded room (6m x 4m x 3m) with appliances, where people or cars transit near it. This noise was equivalent to the mean street noise level (73.4 dBA), and our tests were performed during the morning and the afternoon when the transit is usually higher. By natural setting, we mean that we set the environment of the smart home as natural as possible to real-life use of the BCI. That is, there was no isolated or shielded room; several standard appliances were running in the house, windows and doors opened, and street-level noise at the time of the recording. 

\subsection{Data collection protocol}
The P300 dataset was obtained from four sessions, two per day, for all the participants. Six runs, one per figure,  represent a complete session, where each run follows the protocol based on Hoffmann's work for disabled people \cite{exp_hoff}.



First, the participant is informed about the protocol user was instructed to count the flashes of a specific home appliance in the display, before every experimental block (sequence of stimuli flashes). That appliance that the user was instructed to fixate, was our ‘ground truth’, which we intend to decode from the captured user’s brain signals during the run of an experimental block. This later decoding process is unaware of which appliance the user was instructed to control. Once decoded the system could use this inference to activate a control command over that appliance (e.g. on/off). Subsequently, six images in a 2x3 row-column configuration, see Figure \ref{fig:EEG_BCI_flowchart}, were displayed on the monitor's screen. The experiment starts with one image (out of six), randomly selected, flashed for a period of 100ms, then a white background is displayed during 300ms for all 400ms the EEG signals were recorded, as depicted in Figure \ref{fig:10_20_system} (c). A pseudo-code of the procedure is presented in Algorithm \ref{alg:p300proces}. 

A block is completed after this process repeats for the six different images. Moreover, a block is presented as a block-randomized structure, where the selected image flashed once in six flashes and twice in twelve.  Each run has a total of 20 and 25 blocks chosen randomly.

\begin{algorithm}
	\caption{P300-based Interaction Process}
	\begin{algorithmic}[1]
	\Procedure{Select home item}{$B,N,items$}\Comment{B is buffer, N num. home items, items list}
	    \State {Switch on Interface}
	    \State {$B \leftarrow$ Initiate \ EEG \ buffer}
	    \State {Initiate random seed}
		\For {$iteration=1,2,\ldots N$}
		    \State {$Item \leftarrow$ Randomly select home item from item list}
		    \State {Initiate timer}\Comment{milliseconds counter}
		    \While {timer $<$ $100ms$}
		        \State {Flash home item stimulus}
		    \EndWhile
		    \While {timer $<$ $400ms$}
		        \State {$B[Iteration] \leftarrow$ fill \ buffer \ with \ EEG \ signal}
		    \EndWhile
		\EndFor
		\For {$element=1,2,\ldots,N$ \ in \ B}
			\State {$Input[element] \leftarrow$ run \ signal \ preprocessing \ (from sec. \ref{sec:preprocessing}) \ in \ $B[element]$}
			\State{$Confidence[element] \leftarrow$ \ run \ trained \ DL \ model \ (from sec. \ref{sec:deeplearning}) with \ $Input[element]$ } 
		\EndFor
		\State{$index \leftarrow  \argmax$ $Confidence$} 
		\State{$Prediction \leftarrow items[index]$}\Comment{Prediction is the selected item}
	\EndProcedure	
	 
	\end{algorithmic} 
	 \label{alg:p300proces}
\end{algorithm}


\subsection{Signal Pre-processing}
\label{sec:preprocessing}
First, the signals are downsampled to 120 Hz to reduce computational cost. Then, a sixth-order Butterworth bandpass filter (1-15Hz) is applied \cite{sergio_P300_2}. Moreover, invalid spectral components were eliminated using Notch filters, and the elimination of outlier values followed the winsorization criteria \cite{exp_hoff}.

\subsection{Fuzzy neural block}
In this section, we provide a formal definition of our proposed FNB.

\begin{definition}[Fuzzy neural block]
A fuzzy neural block (FNB) is defined as a sequence of processing layers making up the activation of the antecedents of a fuzzy rule. It first takes the normalised output of the previous layer $\bm{O}$ that is subsequently flattened as $\bm{v_{i}} = Vec(\bm{O_{i}})=[o_{1,1},...,o_{m,1},o_{1,2},...,o_{m,2},...,o_{1,n},...,o_{m,n}]^T$. Next, the fuzzy clustering approach in Kilic et al. \cite{kilicc2007} is used to find a set of $K$ centroids of shape $\bm{c^k}=[c^{k}_1,c^{k}_2,...,c^{k}_d]$, using a collection $\bm{D}$ of $\bm{v_{i}}$ of layer outputs collected from the previous epoch. For the first epoch, the centroids are set to zero. The Gaussian membership value of $\bm{v_{i}}$ is computed as 
\begin{align}
    \bm{\mu}_{i}^{k}(\bm{v}_{i},\bm{c}^{k},\bm{a})=exp\left(-1/4((\bm{v}_{i}-\bm{c}^{k})^{2}/\bm{a}^{2}\right)
\end{align}
the scaling vector $\bm{a}$ is a parameter that is set to learn by the network, and the rule activation consists of a t-norm operator and normalisation step such:
\begin{align} \label{eq:output_FNB}
    o^k(\bm{v}_{i}) = \prod_{j=1}^{d}\mu_{i,j}^{k} \hspace{10mm} and \hspace{10mm} \widetilde{o}^{k}=o^{k}/{\textstyle \sum_{f=1}^{K}o^{f}}
\end{align} 
where $d$ is the dimension of the $\bm{\mu}_{i}^{k}$ vector, and $\bm{O'} = [\widetilde{o}^{1},...,\widetilde{o}^{K}]$ is the output of the FNB that is forwarded to next layer. Note the output dimension of the FNB is reduced to $K$.
\end{definition}

\subsection{List of considered deep architectures for the Smart-Home P300 detection paradigm}
\label{sec:deeplearning}
In addition to bench-marking the state-of-the-art classification methods presented in section \ref{sec:BG}, we report how their performance compares when these methods are appended with our proposed FNB, as shown in Fig. \ref{fig:CM_withFNB}. In the rest of the section, we describe the architecture of these classification methods after been stacked with the proposed FNB.
\input{Figures/CMwithFNB}
\begin{enumerate}[a)]
\item{\emph{LeNet:}} The LeNet architecture is as outlined in Table \ref{tab:lenet_two_networks} and schematized in Fig. \ref{fig:CM_withFNB} a.

\item{\emph{LeNet + FNB:}}  For LeNet appended with FNB, the architecture is identical until the fourth layer in Table \ref{tab:lenet_two_networks}. In particular, $L4$ layer is divided into two sections. The extracted features of the $L4$ layer pass to the FNB, based on gaussian membership functions. A second path is provided for the extracted features of the $L4$ layer, which is connected to the deep network's fully connected (dense) layer. 
The merge operator takes the output of the Fuzzy Neural Block (FNB) and the Fully Connected (FC) Layers and it merges them into a tensor, which is used as an input of the softmax activation function output layer for classification. A general schematic for a classification model with FNB is shown in Fig. \ref{fig:CM_withFNB} b. 
\item{\emph{EEG-TCNet:}} The EEG-TCNet is as outlined in Table \ref{tab:eeg_tcnet_combined} and schematized in Fig. \ref{fig:CM_withFNB} a.
\item{\emph{EEG-TCNet + FNB:}} The EEG-TCNet \cite{ingolfsson2020eeg} maintains the parameters and configurations from the previous subsection, even though now its purpose is to extract temporal characteristics of the EEG until the $\sigma^5 $ layer. Consequently, a fuzzy system \cite{yeganejou2019interpretable} is introduced to classify this information. In this topology, the last layer ($L5$) is directly connected to the FNB with gaussian membership. Simultaneously, the extracted features of the $L5$ layer are connected to the deep network's fully connected (dense) layer. The output from the FNB and the fully connected layer are then merged before passing through the softmax activation function for classification, as illustrated in Fig \ref{fig:CM_withFNB} b. 


\item{\emph{EEG-TCNet + LSTM}:} According to Wang \cite{wang2020temporal}, the EEG-TCNet works as the central feature extractor. 

An LSTM network enhances the temporal properties from the TCNet to increase the efficiency and precision of a P300 response. Combining these networks enhances the extraction of temporal characteristics efficiently, serving as a reinforcement layer to the TCN. Consequently, the TCNet and LSTM are connected sequentially, 
where the TCNet architecture is further explained in Section 3.1.3. Additionally, two LSTM layers for data characterisation with 30 neurons are included, leading to the dense layer with a softmax function.  


\item{\emph{EEG-TCFNet (EEG-TCNet + LSTM + FNB):}} The proposed structure joins the EEG-TCNet, LSTM, and FNB to create a robust classifier for the P300 response. The components are sequentially placed with two routes as output data: the FNB with gaussian membership and a fully-connected layer with a softmax activation function, as presented in Fig. \ref{fig:proposed_network}.
\end{enumerate}

\begin{figure}[!t]
\subfloat[Feature Extraction]{\label{Fig:CNN_FS}\input{Figures/Architecture}}
\subfloat[Classification]{\label{Fig:CNN_classification}\input{Figures/CNN_classification}}
\caption{An illustration of the architecture of the proposed network, namely EEG-TCFNet, which consists of an EEG-Net, temporal convolution network (TCN), long short-term memory (LSTM) and a fuzzy neural block (FNB). (a) The input layer of EEG signals is fed into an EEG-Net, which is a convolutional neural network (CNN) and undertakes a depth-wise convolution (layer 2: L2) followed by a separable convolution (L3). The temporal information in the output from EEG-Net is further analyzed using TCN (L5). (b) The classification part of the proposed network consists of two LSTMs (L6 and L7) and an FNB (L8) with gaussian membership functions. The classification is finally done using softmax (L9), where a binary prediction of whether the participant wants to interact with the decoded device is displayed, such as lamp, in this illustration (see Fig. \ref{fig:EEG_BCI_flowchart} for an illustration of the BCI paradigm). Also, the dimensions of each layer of the EEG-TCNet part of the proposed network are listed in Table \ref{tab:eeg_tcnet_combined}.}
\label{fig:proposed_network}
\end{figure}

\section{Results}

The results are computed with the GoogleColaboratory platform using the Python programming language. The Keras interfaces jointly with an ADAM optimizer (learning rate of 0.0001) provide the final values of accuracy and loss value. The metrics are obtained for six structures: LeNet, LeNet-FNB, EEG-TCNet, EGEG-TCNet-FNB, EEG-TCNet-LSTM, and the proposed model EEG-TCFNet (i.e. EEG-TCNet-LSTM-FNB). 
The evaluation of all deep learning methods was performed with mutually exclusive training, validation, and hold-out tests during the cross-validation procedure. The subject-dependent validation was performed using leave-one-session-out cross-validation where each session out of the four existing represent an independent fold. On the other hand, the subject-independent validation was performed using leave-one-subject-out cross-validation so that all data from an independent subject is hold-out as test set.
	    
 

\subsection{Subject-dependent results}

The accuracy for subject-dependent classification indicates that our proposed network outperforms the precision value and standard deviation in other structures, as shown in Table \ref{tab:acc_subject}. The highest result was achieved in S01 with a precision of $98.6 \pm 0.5$ (Avg. $\pm$ SD). Besides, the values reach higher values than 90\%, except S02, S04, and S07. Despite the results, the first structure (LeNet) still obtained higher accuracies, even $10\%$ higher, despite its higher values in standard deviation. In Table \ref{tab:acc_subject}, the cross-entropy (C-E) values for each subject and structure are presented. The numbers showed performance improvements of our proposed network, resulting in the lowest error value among all the structures. In particular, a healthy subject (S01) and a post-stroke patient (S09) indicated values lower than 0.15. Regardless of having higher accuracy, the LeNet structure shows a higher standard deviation, representing uneven results for each fold. Also, LeNet presents inconsistent results due to high C-E values compared to our proposed structure, reaching values fewer than 0.5 in some cases.  


\begin{table}[h]
    \centering
    \caption{A comparison of the classification prowess of the models with and without the proposed fuzzy neural block (FNB). The average (Avg.) values of accuracy (Accu.) with standard deviation (SD) as well as cross-entropy (C-E) are reported for each classifier for subject-dependent classification.}
    \scalebox{.76}{
    \begin{tabular}{l l l l l l l l l l l l l l l l l l l l l} \hline
    \textbf{} & \multirow{2}{*}{\textbf{FNB}} & \textbf{}& \multicolumn{2}{c}{\textbf{S01}} & \multicolumn{2}{c}{\textbf{S02}} & \multicolumn{2}{c}{\textbf{S03}} & \multicolumn{2}{c}{\textbf{S04}} & \multicolumn{2}{c}{\textbf{S05}} & \multicolumn{2}{c}{\textbf{S06}} & \multicolumn{2}{c}{\textbf{S07}} & \multicolumn{2}{c}{\textbf{S08}} & \multicolumn{2}{c}{\textbf{S09}}\\ 
    & & & \emph{Avg.} & \emph{SD} & \emph{Avg.} & \emph{SD} & \emph{Avg.} & \emph{SD} & \emph{Avg.} & \emph{SD }& \emph{Avg.} & \emph{SD} & \emph{Avg.} & \emph{SD} & \emph{Avg.} &\emph{ SD} & \emph{Avg.} & \emph{SD} & \emph{Avg.} & \emph{SD} \\ \hline
        \multirow{4}{*}{\rotatebox[origin=c]{90}{LeNet}} &  No & \multirow{2}{*}{\rotatebox[origin=c]{90}{Accu.}}& \cellcolor{gray!10}95.5 & \cellcolor{gray!10}3.7 & \cellcolor{gray!10} 88.5 & \cellcolor{gray!10}10.3& \cellcolor{gray!10}92.0 &\cellcolor{gray!10} 8.8& \cellcolor{gray!10}86.8 &\cellcolor{gray!10} 7.2& \cellcolor{gray!10}92.5 &\cellcolor{gray!10} 6.1 & \cellcolor{gray!10}92.5& \cellcolor{gray!10}6.6& \cellcolor{gray!10}84.1 &\cellcolor{gray!10} 14.3& \cellcolor{gray!10}89.8 & \cellcolor{gray!10}9.1& \cellcolor{gray!10}93.4 & \cellcolor{gray!10}7.3 \\
         &  Yes & & \cellcolor{gray!10}96.0 & \cellcolor{gray!10}3.1 &   \cellcolor{gray!10}92.2 & \cellcolor{gray!10}9.5 &   \cellcolor{gray!10}93.1 &\cellcolor{gray!10} 7.4 &   \cellcolor{gray!10}87.8 & \cellcolor{gray!10}9.1 &   \cellcolor{gray!10}93.1 & \cellcolor{gray!10}6.6 &   \cellcolor{gray!10}94.7 & \cellcolor{gray!10}6.3 &  \cellcolor{gray!10}84.8  & \cellcolor{gray!10} 13.1  &  \cellcolor{gray!10}92.4 & \cellcolor{gray!10}7.0  &   \cellcolor{gray!10}94.3 & \cellcolor{gray!10}5.7  \\ \cline{2-21}
          &  No & \multirow{2}{*}{\rotatebox[origin=c]{90}{C-E}}& 0.23 & 0.22 &  0.56 & 0.58 &  0.39 & 0.43 &  0.76 & 0.46 &  0.44 & 0.37 &  0.45 & 0.43 &  0.83 & 0.85 &  0.55 & 0.55 &  0.33 & 0.42 \\
         &  Yes & & 0.18 & 0.17 &  1.22 & 1.54 &  0.34 & 0.46 &  1.75 & 2.45 &  0.43 & 0.48 &  0.30 & 0.39 &  0.80 & 0.92 &  0.46 & 0.49 &  0.36 & 0.47   \\ \hline
        \multirow{4}{*}{\rotatebox[origin=c]{90}{TCNet}} &  No& \multirow{2}{*}{\rotatebox[origin=c]{90}{Accu.}} &\cellcolor{gray!10}97.9 & \cellcolor{gray!10}1.3 &   \cellcolor{gray!10}82.2 & \cellcolor{gray!10}2.9 &   \cellcolor{gray!10}91.0 & \cellcolor{gray!10}3.5 &   \cellcolor{gray!10}78.4 & \cellcolor{gray!10}7.7 &   \cellcolor{gray!10}93.4 & \cellcolor{gray!10}2.2 &  \cellcolor{gray!10}94.9 &\cellcolor{gray!10} 2.8  &    \cellcolor{gray!10}78.1 &\cellcolor{gray!10} 3.1 &   \cellcolor{gray!10}95.8 & \cellcolor{gray!10}1.5 &  \cellcolor{gray!10}93.8 &\cellcolor{gray!10} 1.5  \\
         &  Yes& & \cellcolor{gray!10}98.1& \cellcolor{gray!10}1.7 &   \cellcolor{gray!10}84.2 &\cellcolor{gray!10} 1.9 &   \cellcolor{gray!10}90.8 & \cellcolor{gray!10}4.1 &   \cellcolor{gray!10}81.0 & \cellcolor{gray!10}5.4 &   \cellcolor{gray!10}94.4 & \cellcolor{gray!10}1.1 &   \cellcolor{gray!10}95.1 & \cellcolor{gray!10}2.0 &   \cellcolor{gray!10}79.5 & \cellcolor{gray!10}3.1  &   \cellcolor{gray!10}96.1 & \cellcolor{gray!10}1.8 &   \cellcolor{gray!10}94.4 &\cellcolor{gray!10} 2.0\\ \cline{2-21}
          &  No& \multirow{2}{*}{\rotatebox[origin=c]{90}{C-E}} & 0.06 & 0.04 &  0.42  & 0.04 &  0.25 & 0.08 &  0.47 & 0.06 &  0.19  & 0.03 &  0.15 & 0.09 &  0.49 & 0.07 &  0.12 & 0.04 &  0.16  & 0.05 \\
         &  Yes& & 0.12 & 0.11 &  0.39 & 0.03 &  0.23 & 0.1 &  0.45 & 0.04 &  0.16 &0.04 &  0.13 & 0.05 &  0.50 & 0.09 &  0.12 & 0.05 &  0.14 & 0.07\\ \hline
        \multirow{1}{*}{} &  No & \multirow{2}{*}{\rotatebox[origin=c]{90}{Accu.}} &\cellcolor{gray!10}98.2 & \cellcolor{gray!10}1.5 &   \cellcolor{gray!10}84.7 & \cellcolor{gray!10}3.9 &   \cellcolor{gray!10}91.5 & \cellcolor{gray!10}3.8 &   \cellcolor{gray!10}79.3 & \cellcolor{gray!10}5.7 &   \cellcolor{gray!10}94.1 & \cellcolor{gray!10}2.3 &   \cellcolor{gray!10}95.0 & \cellcolor{gray!10}1.2 &    \cellcolor{gray!10}78.8 &\cellcolor{gray!10} 4.5 &  \cellcolor{gray!10}95.7 &\cellcolor{gray!10} 1.4  &   \cellcolor{gray!10}94.2 & \cellcolor{gray!10}1.9 \\
        \raisebox{.95\normalbaselineskip}[0pt][0pt]{ \multirow{1}{*}{\rotatebox[origin=c]{90}{TCNet-LSTM}}}&  Yes & & \cellcolor{gray!10}98.6 & \cellcolor{gray!10}0.5  &   \cellcolor{gray!10}86.1 & \cellcolor{gray!10}2.5 &   \cellcolor{gray!10}92.3 & \cellcolor{gray!10}2.1 &   \cellcolor{gray!10}83.5 &\cellcolor{gray!10} 2.8 &   \cellcolor{gray!10}94.9 & \cellcolor{gray!10}3.9 &   \cellcolor{gray!10}95.9 & \cellcolor{gray!10}1.5 &    \cellcolor{gray!10}78.9 & \cellcolor{gray!10} 4.2 &   \cellcolor{gray!10}96.1 &\cellcolor{gray!10} 1.1 &   \cellcolor{gray!10}94.4 & \cellcolor{gray!10}1.6 \\ \cline{2-21}
           \multirow{1}{*}{\rotatebox[origin=c]{90}{}} &  No & \multirow{2}{*}{\rotatebox[origin=c]{90}{C-E}} &0.0 & 0.04 &  0.37 & 0.09 &  0.23 & 0.09 & 0.45 & 0.05 &  0.17 & 0.06 &  0.13 & 0.04 & 0.52 & 0.14 &  0.13 & 0.050 &  0.15 & 0.06 \\
         \multirow{1}{*}{}&  Yes & & 0.03 & 0.01 &  0.37 & 0.07 &  0.23 & 0.04 &  0.39 & 0.03 &  0.15 & 0.10 &  0.12 & 0.04 &  0.47 & 0.06 &  0.09 & 0.03 &  0.15 & 0.05 \\\hline
    \end{tabular}}
    \label{tab:acc_subject}
\end{table}



\begin{figure}[h]
	\centering
	\includegraphics[trim= {3cm, 0cm, 3cm, 0cm}, clip=true, scale=0.55]{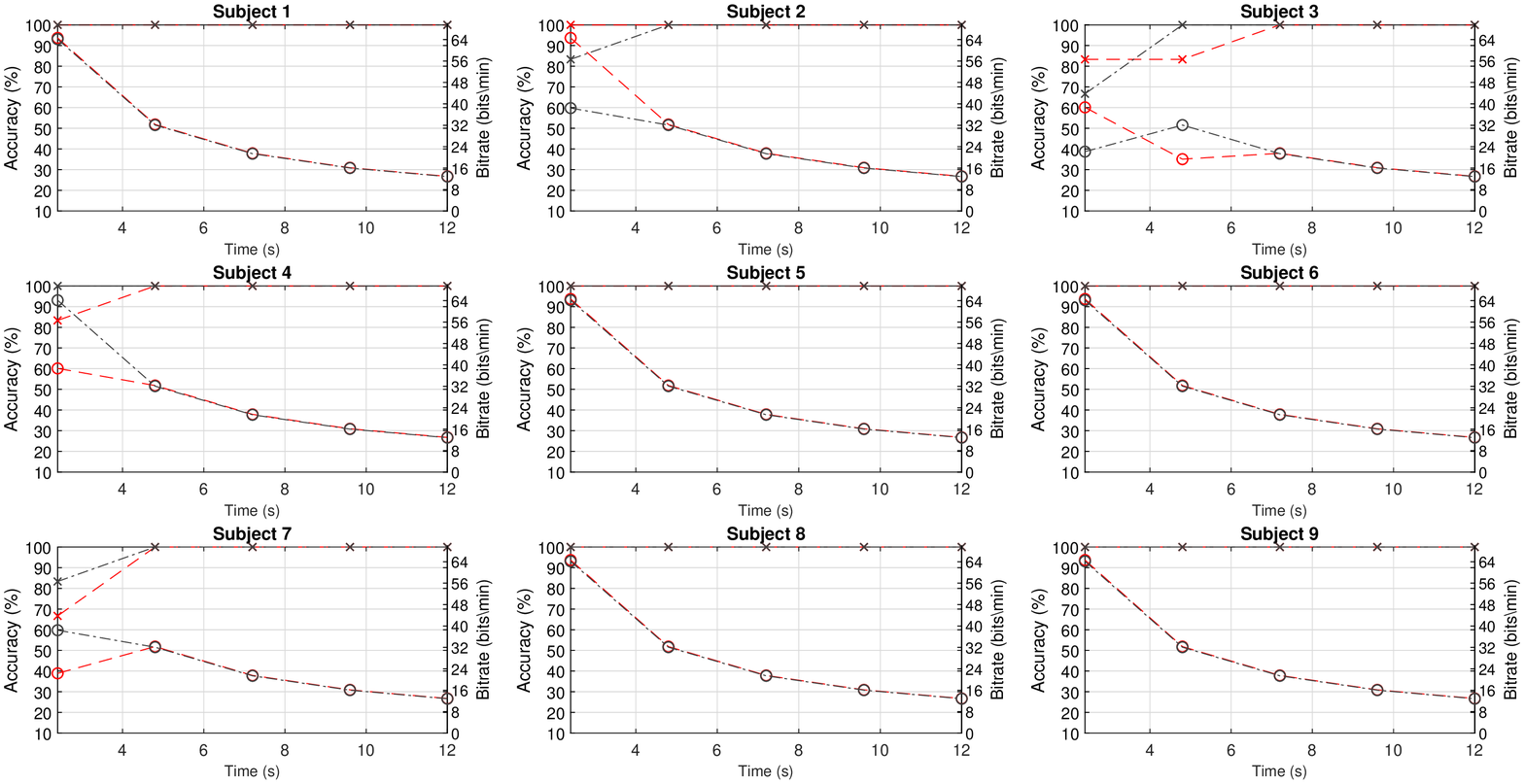}
	\caption{Classification accuracy and bitrate plotted vs. time for Subject-dependent strategy. The panels show the classification accuracy obtained with EEG-TCNet-LSTM 
averaged over four sessions (red crosses) and bitrate (red circles). Also to  EEG-TCNet-LSTM-FNB averaged over four sessions (black crosses) and bitrate (black circles) for disabled subjects (S7-S9) and able-bodied subjects (S1-S6).}
	\label{fig:BitRate} 
\end{figure}

Furthermore, using an evaluation methodology called target by block strategy reported in \cite{exp_hoff} we calculate a classification accuracy and bitrate for two topologies EEG-TCN-LSTM and EEG-TCN-LSTM-FNB shown in Fig. \ref{fig:BitRate}. These results indicate that EEG-TCN-LSTM-FNB outperforms EEG-TCN-LSTM for subjects 3, 4 and 7. 
These results are important as it demonstrates the proper operation capacity of BCI in real-time.

\subsection{Subject-independent results}
On the other hand, the subject-independent classification resulted in a table of accuracies with their standard deviations, as shown in Table \ref{tab:ACU_value_cross_bysubej}.  Specifically, the proposed structure has the highest accuracy of $74.3 \pm 1.7\%$ in S01 and the lowest of $59.6 \pm 2.1\%$ in S06. Consequently, LeNet networks (with and without FNB) outperform our network in all the subjects, whereas the EEG-TCNet structures vary depending on the evaluated subject. In certain situations, the standard deviation values interest with the range of values of different networks, for example, the comparison between network LeNet-FNB and EEG-TCFNet for the subject S03. 

\begin{table}[!h]
    \centering
    \caption{A comparison of the classification prowess of the models with and without the proposed fuzzy neural block (FNB). The average (Avg.) and standard deviation (SD) values of accuracy (Accu.) and cross-entropy (C-E) are reported for each classifier for subject-independent classification.}
    \scalebox{1}{
    \begin{tabular}{l l l l l l l l l l l l l l l} \hline
    \multirow{2}{*}{\textbf{Model}} & \multirow{2}{*}{\textbf{FNB}} & \textbf{}& \multicolumn{2}{c}{\textbf{S01}} & \multicolumn{2}{c}{\textbf{S02}} & \multicolumn{2}{c}{\textbf{S03}} & \multicolumn{2}{c}{\textbf{S04}} & \multicolumn{2}{c}{\textbf{S05}} & \multicolumn{2}{c}{\textbf{S06}} \\ 
    & & & \emph{Avg.} & \emph{SD}& \emph{Avg.} & \emph{SD}& \emph{Avg.} & \emph{SD}& \emph{Avg.} & \emph{SD}& \emph{Avg.} & \emph{SD}& \emph{Avg.} & \emph{SD}\\\hline
        \multirow{4}{*}{LeNet} &  No & \multirow{2}{*}{\rotatebox[origin=c]{90}{Accu.}}&\cellcolor{gray!10}85.1 & \cellcolor{gray!10}1.3 &   \cellcolor{gray!10}76.2 &  \cellcolor{gray!10}2.0 &   \cellcolor{gray!10}77.8 &  \cellcolor{gray!10}2.4 &   \cellcolor{gray!10}75.1 & \cellcolor{gray!10}2.4 &  \cellcolor{gray!10}76.6 & \cellcolor{gray!10}3.3  &   \cellcolor{gray!10}80.2 & \cellcolor{gray!10}1.7 \\
         &  Yes & &    \cellcolor{gray!10}84.4 & \cellcolor{gray!10}2.3 &   \cellcolor{gray!10}77.3 & \cellcolor{gray!10}2.1 &   \cellcolor{gray!10}77.3 & \cellcolor{gray!10}2.0 &   \cellcolor{gray!10}75.8 & \cellcolor{gray!10} 1.6 &   \cellcolor{gray!10}73.3 & \cellcolor{gray!10}1.8 &   \cellcolor{gray!10}79.0 & \cellcolor{gray!10}3.3 \\ \cline{2-15}
          &  No & \multirow{2}{*}{\rotatebox[origin=c]{90}{C-E}}& 1.59 & 0.25 &   6.63& 2.85 &  1.92 & 0.43 &   3.36 & 0.44 &  2.90 & 1.04 &  1.46 & 0.30 \\
         &  Yes & & 1.55& 0.50 &  19.47 & 28.58 &  6.12 & 7.38 &  34.27& 50.32 &  3.05& 1.23 &  1.04& 0.32 \\ \hline
        \multirow{4}{*}{TCNet} &  No& \multirow{2}{*}{\rotatebox[origin=c]{90}{Accu.}} &\cellcolor{gray!10}68.0& \cellcolor{gray!10}4.3 &   \cellcolor{gray!10}67.1& \cellcolor{gray!10}4.3 &   \cellcolor{gray!10}73.7& \cellcolor{gray!10}2.2 &   \cellcolor{gray!10}58.5& \cellcolor{gray!10}5.2 &   \cellcolor{gray!10}74.3& \cellcolor{gray!10}2.5 &   \cellcolor{gray!10}53.8& \cellcolor{gray!10}1.9  \\
         &  Yes& & \cellcolor{gray!10}74.7& \cellcolor{gray!10}3.9 &   \cellcolor{gray!10}69.1& \cellcolor{gray!10}2.3 &   \cellcolor{gray!10}78.1& \cellcolor{gray!10}1.8 &   \cellcolor{gray!10}59.5& \cellcolor{gray!10}3.2 &   \cellcolor{gray!10}76.6& \cellcolor{gray!10}2.7 &   \cellcolor{gray!10}56.2& \cellcolor{gray!10}2.9 \\ \cline{2-15}
          &  No& \multirow{2}{*}{\rotatebox[origin=c]{90}{C-E}} & 0.57& 0.05 &   0.62& 0.06 &  0.53& 0.02 &   0.88& 0.12 &  0.47& 0.03 &  0.96& 0.13  \\
         &  Yes& & 0.49& 0.04 &   0.65& 0.05 &  0.47& 0.03 &   1.01& 0.29 & 0.46& 0.02 &  0.86& 0.11\\ \hline
        \multirow{1}{*}{} &  No & \multirow{2}{*}{\rotatebox[origin=c]{90}{Accu.}} & \cellcolor{gray!10}70.7& \cellcolor{gray!10}6.1 &   \cellcolor{gray!10}68.8& \cellcolor{gray!10}2.8 &   \cellcolor{gray!10}72.9& \cellcolor{gray!10}4.1 &   \cellcolor{gray!10}60.6& \cellcolor{gray!10}1.5 &   \cellcolor{gray!10}73.9& \cellcolor{gray!10}3.5 &   \cellcolor{gray!10}54.3& \cellcolor{gray!10}2.0\\
         \multirow{1}{*}{TCNet-}&  Yes & &  \cellcolor{gray!10}74.3& \cellcolor{gray!10}1.7 &   \cellcolor{gray!10}70.3& \cellcolor{gray!10}2.6 &   \cellcolor{gray!10}72.3& \cellcolor{gray!10}5.5 &   \cellcolor{gray!10}60.3& \cellcolor{gray!10}2.3 &   \cellcolor{gray!10}72.3& \cellcolor{gray!10}3.0 &   \cellcolor{gray!10}59.6& \cellcolor{gray!10}2.1  \\ \cline{2-15}
           \multirow{1}{*}{LSTM} &  No & \multirow{2}{*}{\rotatebox[origin=c]{90}{C-E}} &0.51& 0.07 &   0.62& 0.05 &  0.54& 0.04 &   0.80& 0.05 &  0.49& 0.03 &  0.95& 0.03 \\
         \multirow{1}{*}{}&  Yes & &  0.48& 0.03 &   0.62& 0.04 &  0.55& 0.08 &   0.78& 0.03 &  0.53& 0.04 &  0.78& 0.08   \\\hline
    \end{tabular}}
    \label{tab:ACU_value_cross_bysubej}
\end{table}



\begin{figure}[h]
	\centering
	\includegraphics[trim= {3cm, 0cm, 3cm, 0cm}, clip=true, scale=0.55]{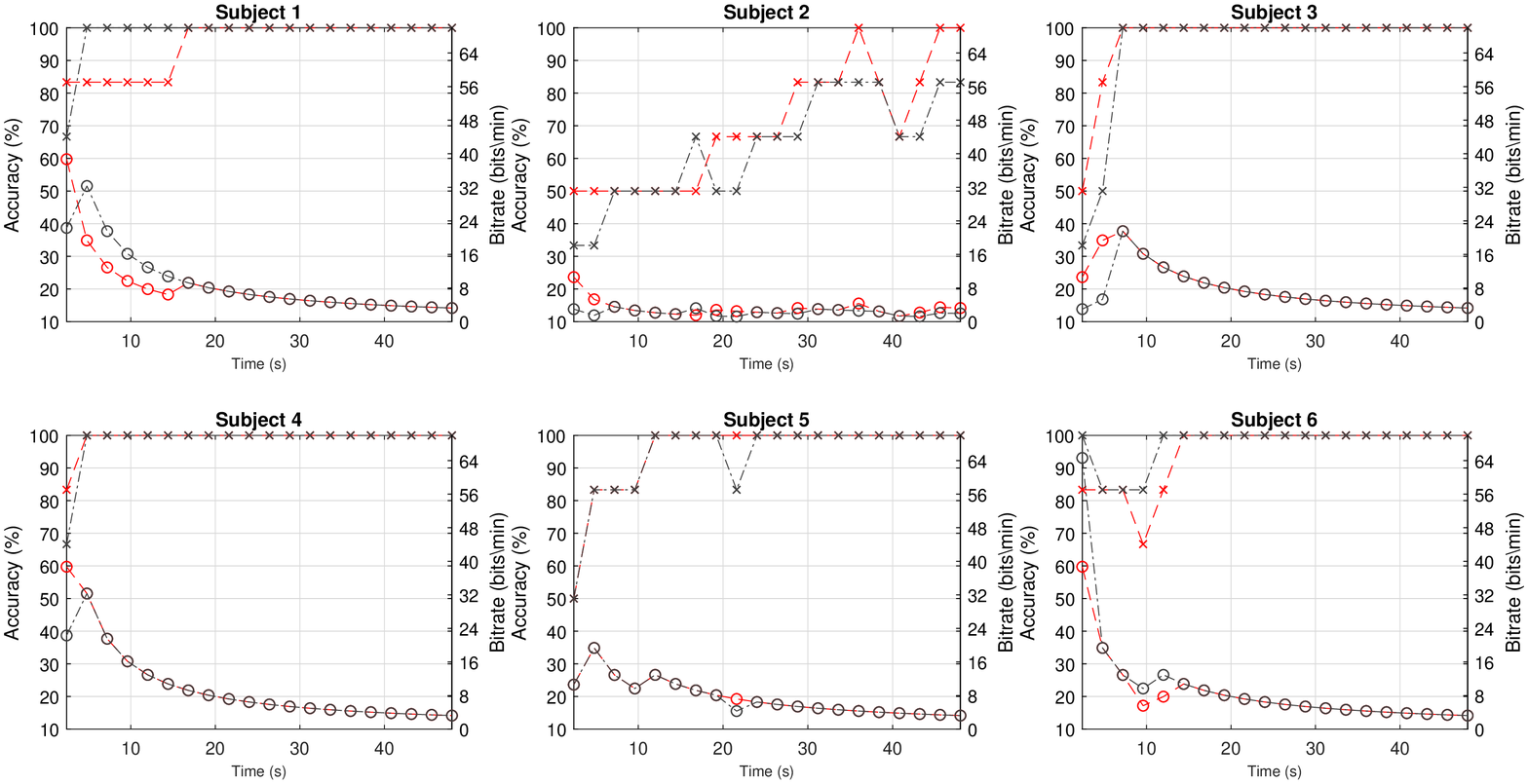}
	\caption{Classification accuracy and bitrate plotted vs. time for Subject-independent strategy. The panels show the classification accuracy obtained with EEG-TCNet-LSTM 
averaged over four sessions (red crosses) and bitrate (red circles). Also to  EEG-TCNet-LSTM- FNB averaged over four sessions (black crosses) and bitrate (black circles).
For able-bodied subjects (S1-S6).}
	\label{fig:BitRat_SI} 
\end{figure}

In the same way that subject-dependent, we calculate a classification accuracy and bitrate for two topologies EEG-TCNet-LSTM and EEG-TCNet-LSTM-FBN, as
we can see in figure \ref{fig:BitRat_SI}. Our results reported that EEG-TC-LSTM-Fuzzy outperformed EEG-TCN-LSTM to subject 1, 5, and 6. However, it is essential to state that subjects 2 does not report outstanding accuracy in both EEG-TCNet-LSTM and EEG-TCNet-LSTM-FNB as they do not reach $100\%$ accuracy. Overall, subjects 1, 3, 4, 5, 6 reached $100\%$ accuracy in less than 12 seconds.  

\subsection{Statistical performance comparison of fuzzy methods}
Table \ref{table:statistic}, reported a comparing average accuracy, standard deviation, and significant difference (p-value) for each topology and strategy classification. The Kolmogorov–Smirnov test was performed to analyze the normality of average accuracies of topologies classification, and after that, a statistical test was performed to analyze the statistical differences of impact of FNB.

The subject-dependent strategy reported a higher classification accuracy of 
$91.9 \pm 3.5$ (Avg. $\pm$ SD) to LeNet-FNB, and overall, the use of FNB in the three topologies outperformed the topologies that did not include FNB. Likewise, the subject-independent strategy reported a $78.5 \pm 3.7$ (Avg.$\pm$ SD) classification accuracy, higher than LeNet. The use of FNB improves the accuracy classification in EEG-TCNet-FNB and EEG-TCNet-LSTM-FNB. The accuracy values show a non-normal distribution. Thus, a non-parametric Wilcoxon signed-rank test was performed to analyze the median of differences between each topologies’ accuracies which reported p-values $(p<0.05)$ in Table \ref{table:statistic} that indicate it can be considered statistically significant.



%

\begin{table}[ht]
	\centering
	\caption{A comparison of average accuracy, standard deviation, and significant difference (p-value) from the Wilcoxon signed-rank test for each topology and strategy classification.}
		\begin{tabular}
		{>{\centering\arraybackslash}m{3.0cm}
		>{\arraybackslash}m{6.9cm} >{\centering\arraybackslash}m{1.7cm} >{\centering\arraybackslash}m{1.4cm} >{\centering\arraybackslash}m{1.4cm}} 
			\toprule
            
			\textbf{Strategy  classification} & 
			\textbf{Topology classification} &
			\textbf{Avg. Acc. (\%)} &  \textbf{Std. Acc.(\%)} &  \textbf{p-value}  \\  \midrule
			
			\multirow{6}{*}{Subject dependent}
		     &LeNet  &  90.6  & 3.6 & \multirow{2}{*}{0.0078} \\
		     &LeNet-FNB & 91.9 & 3.5  \\\cline{2-5}  
             &EEG-TCNet  & 89.5  & 7.8 & \multirow{2}{*}{0.0117} \\
		     &EEG-TCNet-FNB  & 90.4 & 7.0  \\\cline{2-5}  
             &EEG-TCNet-LSTM  & 90.2  & 7.3 & \multirow{2}{*}{0.0039} \\
		     &EEG-TCNet-LSTM-FNB(i.e.EEG-TCFNet)& 91.2 & 6.7  \\\cline{1-5}  
            \multirow{6}{*}{Subject independent}
              &LeNet  & 78.5  & 3.7 & \multirow{2}{*}{0.4688} \\
		     &LeNet-FNB & 77.9 & 3.7  \\\cline{2-5}  
             &EEG-TCNet  & 65.9  & 8.2 & \multirow{2}{*}{0.0313} \\
		     &EEG-TCNet-FNB  & 69.0 & 9.2  \\\cline{2-5}  
             &EEG-TCNet-LSTM  & 66.9  & 7.8 & \multirow{2}{*}{0.5625} \\
		     &EEG-TCNet-LSTM-FNB(i.e.EEG-TCFNet) & 68.2 & 6.5 \\\cline{1-5}  
			
	\end{tabular}
	
	\label{table:statistic}
\end{table}

\subsection{Discussion on results for Healthy vs. Stroke patients}
Concerning the patient cohort and EEG-TCNet-LSTM-FNB, subjects S08 achieved an accuracy higher than other disabled subjects; however, subject S07 performance was the lowest among all.
The results obtained for subject 07 might be related and caused by its critical medical condition. 
It is well-known that concentration can affect the P300 waveform as it is a visual-induced endogenous brain response. So P300 waveform can be elicited without problems in post-stroke victims; nevertheless, the subject’s attention plays a vital role in evoking it. 
Subject S07 probably has a short attention span consequence of the stroke, which made its concentration on the experiment diminishes over time. Nevertheless, both subjects S08 and S09 results indicate using a BCI-based P300 for ischemic post-stroke victims employing these classification pipelines.

In Table \ref{tab:ACU_value_cross_bysubej}, the C-E values are presented for subject-independent classification.  The Lenet structures (with and without FNB) contain mean-squared errors above 1, representing instability in the network. Nonetheless, EEG-TCNet architectures (EEG-TCNet, EEG-TCNet-FNB, EEG-TCNet-LSTM) have stable convergence during their training since they achieved C-E lower than 1, except in EEG-TCNet on S04. Lastly, our proposed topology, EEG-TCFNet, has all errors below 1 with the lowest value of 0.48 in S01. Even though the LeNet architectures and EEG-TCNet produced higher accuracies than EEG-TCFNet, the C-E indicates higher stability for our proposed network, EEG-TCFNet, in all the cases. Specifically, exceptions in S03 and S06 were recognized, where EEG-TCFNet provided lower C-E values with lower accuracies as well.

\subsection{Results comparison with previous works}
Table \ref{tab:DBI_C} shows the classification accuracy obtained in previous works \cite{cortez2021smart, sergio_P300_1, sergio_P300_2, cortez2020under} using the same dataset. These previous works were divided whether they include an undersampling method and the applied classifier, including Support Vector Machines (SVM), Multilayer Perceptron (MLP), Deep Belief Network (DBN), and Deep Feed-forward Network (DFN).  From the results, our proposed structure, EEG-TCFNet, reached higher average accuracies in most of the subjects with the highest value of $98.6\%$ in S01. Nevertheless, subject S04, with an accuracy of $83.5\%$, presented lower values than DBN and DFN with an accuracy of $87.4\%$ and $84.5\%$, respectively; subject S09 obtained a $94.4\%$ compared to the DBN result with $94.9\%$.
Some advantages of EEG-TCFNet is that it achieved the highest precision in classification with $98.6\%$ for S01 over our previous works and a better bit-rate for the S04 over EEG-TCNet + LSTM. In summary, we reported that the classification accuracy of EEG-TCFNet outperformed our previous works for eight out of nine subjects. Yet, a disadvantage of deep learning models is a greater computational complexity compared to a machine learning method such as Bayesian Linear Discriminant Analysis (BLDA). Nevertheless, the suggested deep learning architectures are light enough to be run in a wide range of mainstream microprocessors and current edge systems. Moreover, this pipeline classification is suitable to BCI-based P300 in non-optimal environments to operate in real-life conditions.
\begin{table}[h]
\caption{Accuracy comparison of the proposed model, EEG-TCFNet, with previous work with and without under-sampling.}
	\footnotesize
	\centering
    \small
    \setlength\tabcolsep{2pt}
\begin{tabular*}{\linewidth}{@{\extracolsep{\fill}} l c ccc cccc @{}}
    \toprule \textbf{S.}
&  
& \multicolumn{3}{c}{\textbf{Without Under Sampling}} 
& \multicolumn{4}{c}{\textbf{Under sampling}} \\

      \cmidrule{3-5}  \cmidrule{6-9} 
 & \textbf{EEG-TCNet}
 & \textbf{SVM\cite{cortez2021smart}} & 
\textbf{MLP\cite{cortez2021smart}} & \textbf{DBN\cite{sergio_P300_2}}&  
\textbf{DFN\cite{sergio_P300_1}} & \textbf{DBN\cite{sergio_P300_1}} 
& \textbf{DFN\cite{cortez2020under}} &  \textbf{DBN\cite{cortez2020under}}\\
        \midrule
      S01  & 98.6  & 91.5 & 91.8 & 91.6  & 96.3  & 95.5 & 93.6 & 90.1  \\
      S02  & 86.1  & 79.1 & 80.3 & 80.7  & 84.1  & 84.8 & 81.4 & 83.5  \\
      S03  & 92.3  & 83.9 & 85.3 & 85.8  & 91.2  & 91.4 & 92.1 & 86.3  \\
      S04  & 83.5  & 78.9 & 75.7 & 81.7  & 84.5  & 87.4 & 83.2 & 81.0  \\
      S05  & 94.9  & 83.4 & 84.9 & 83.5  & 89.9  & 87.4 & 90.1 & 84.2  \\
      S06  & 95.9  & 81.7 & 83.0 & 82.5  & 88.1  & 84.5 & 84.8 & 81.4  \\
      S07  & 78.9  & 69.2 & 68.6 & 66.6  & 73.7  & 77.6 & 77.7 & 72.0  \\
      S08  & 96.1  & 85.5 & 89.6 & 88.1  & 91.1  & 93.7 & 93.3 & 88.4  \\
      S09  & 94.4  & 87.4 & 86.9 & 85.8  & 93.8  & 94.9 & 86.6 & 84.0  \\
\bottomrule
\end{tabular*}
\label{tab:DBI_C} 
\end{table}



\section{Conclusions and Future Work}
This paper presents a classification model using the TCN in tandem with an LSTM network and an FNB called EEG-TCFNet. When the P300 BCI is performed in real settings and natural surroundings, such as the case of a smart home scenario, it is beneficial to add a fuzzy component to the deep learning architecture, as it may help to cope with a higher parametric uncertainty of the model.
The proposed model EEG-TCFNet combines the modified CNN and LSTM for feature extraction sequentially, then an FNB for target response classification. Experiments were conducted in six healthy and three post-stroke subjects, resulting in favorable indicators for our proposed structure compared to similar topologies, such as LeNet, LeNet-FNB, EEG-TCNet, EEG-TCNet-FNB, EEG-TCNet-LSTM. The EEG-TCFNet (i.e. EEG-TCNet-LSTM-FNB), the proposed  model, obtained an average accuracy of $91.2\%$, with the highest value of $98.6\%$ to subject-dependent classification.
Similarly, subject-independent classification presents an average accuracy of $68.2\%$, reaching the highest value of $74.3\%$.  Also, our results based on subject-dependent classification reached a maximum bitrate of $64$ $bits\min$ and $100\%$ of accuracy using a target by block strategy. It demonstrates the viability of applying BCI in real settings and natural surroundings. Including our previous works, this paper presents subject-independent classification  results for the first time.
In future work, the proposed model could be evaluated with different mental tasks, such as motor imagery. Alternative input data strategies can be studied, for instance, to deal with applications where a class imbalance is unavoidable. Extended longitudinal studies could be planed to allow for a higher balanced population sampling, e.g. regarding their sex characteristics and health conditions.
\section*{Data availability statement}
The data that support the findings of this study are available upon reasonable request from the corresponding author. The data from this study are not publicly available due to containing biometric information that could compromise the privacy of research participants. 

\section*{Authors Contribution}
\textbf{Christian Flores Vega}: Conception, Design of the study, Data collection, Methodology, Results, Writing- Original draft preparation. \textbf{Joaquin Quevedo}: Methodology, Results, Writing. \textbf{Elmer Escandón}: Results, Writing. \textbf{Mehrin Kiani}: Visualization, Writing. \textbf{Weiping Ding}:  Appraised the work, Provided feedback. \textbf{Javier Andreu-Perez}: Conception, Methodology,  Writing.

%


\bibliographystyle{elsarticle-num} 
\bibliography{cas-refs}
%
\end{document}

%% file: Figures/BCI_schematic.tex
\tikzstyle{decision} = [diamond, draw, fill=blue!20, 
    text width=4.5em, text badly centered, node distance=3cm, inner sep=0pt]
\tikzstyle{block} = [rectangle, draw,  
    text width=2.8cm, text centered, rounded corners, minimum height=1.5cm, fill= blue!30]
\tikzstyle{line} = [draw, -latex']
\tikzstyle{cloud} = [draw, ellipse,fill=red!20, node distance=3cm,
    minimum height=2em]
\tikzstyle{dotted_block} = [draw=black!30!white, line width=1pt, dash pattern=on 1pt off 4pt on 6pt off 4pt, inner ysep=1mm,inner xsep=1mm, rectangle, rounded corners ]
\tikzstyle{myarrows} = [draw=black,solid,line width=.5mm, ->]
\tikzstyle{dashedline} = [draw=black,dashed,line width=.5mm]
\begin{tikzpicture} [node distance = 2.5cm]
    \node (brain_sig) [block, fill= none,  label=above:EEG signals,text width = 5cm,] {\includegraphics[scale=0.25, trim = {1.8cm 0cm 1cm 0cm}, clip]{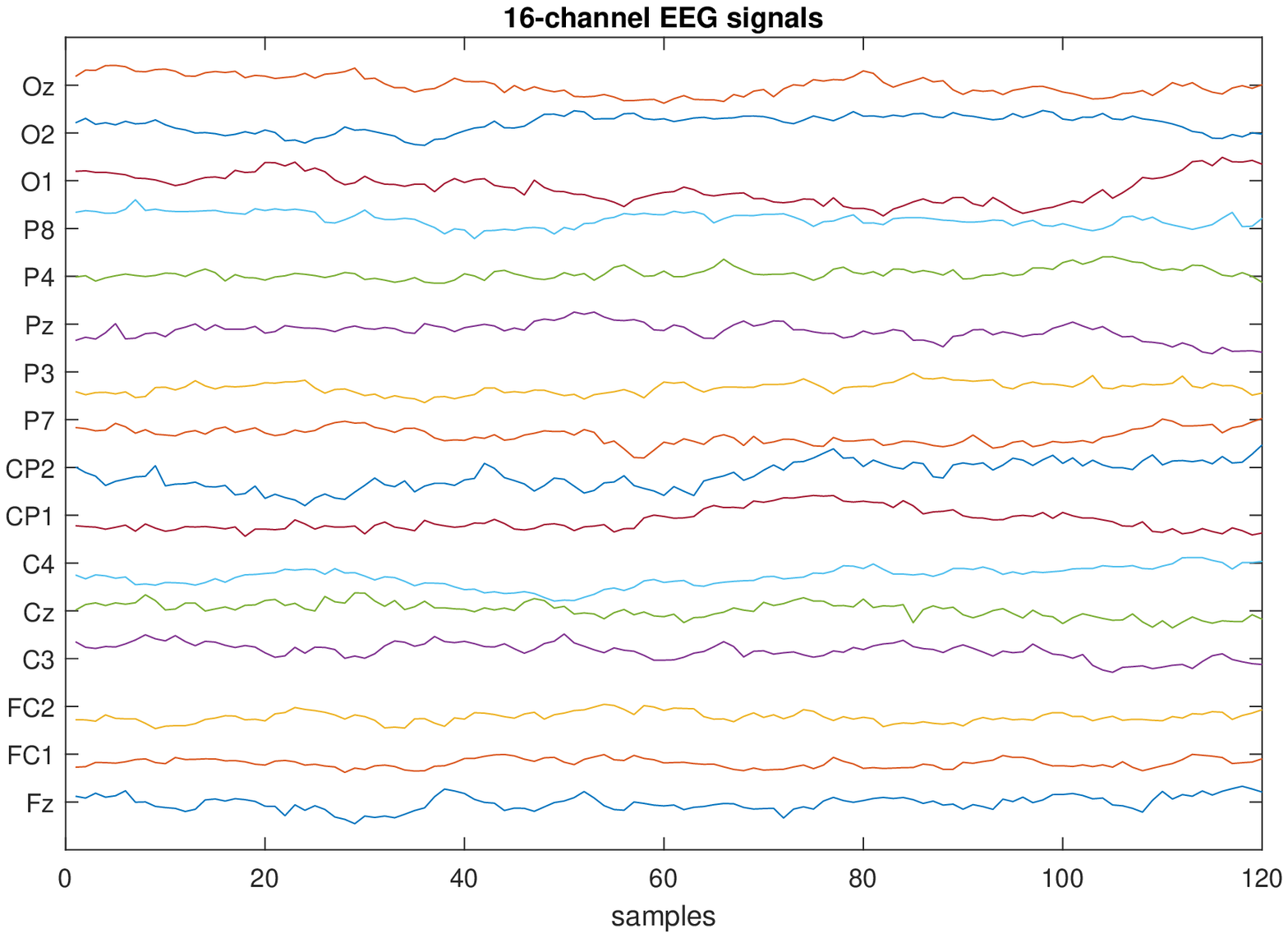}};
    \node (participant) [block, fill= none,  label=above:Participant,above of = brain_sig, text width = 5cm, node distance = 4cm] {\includegraphics[scale=0.04]{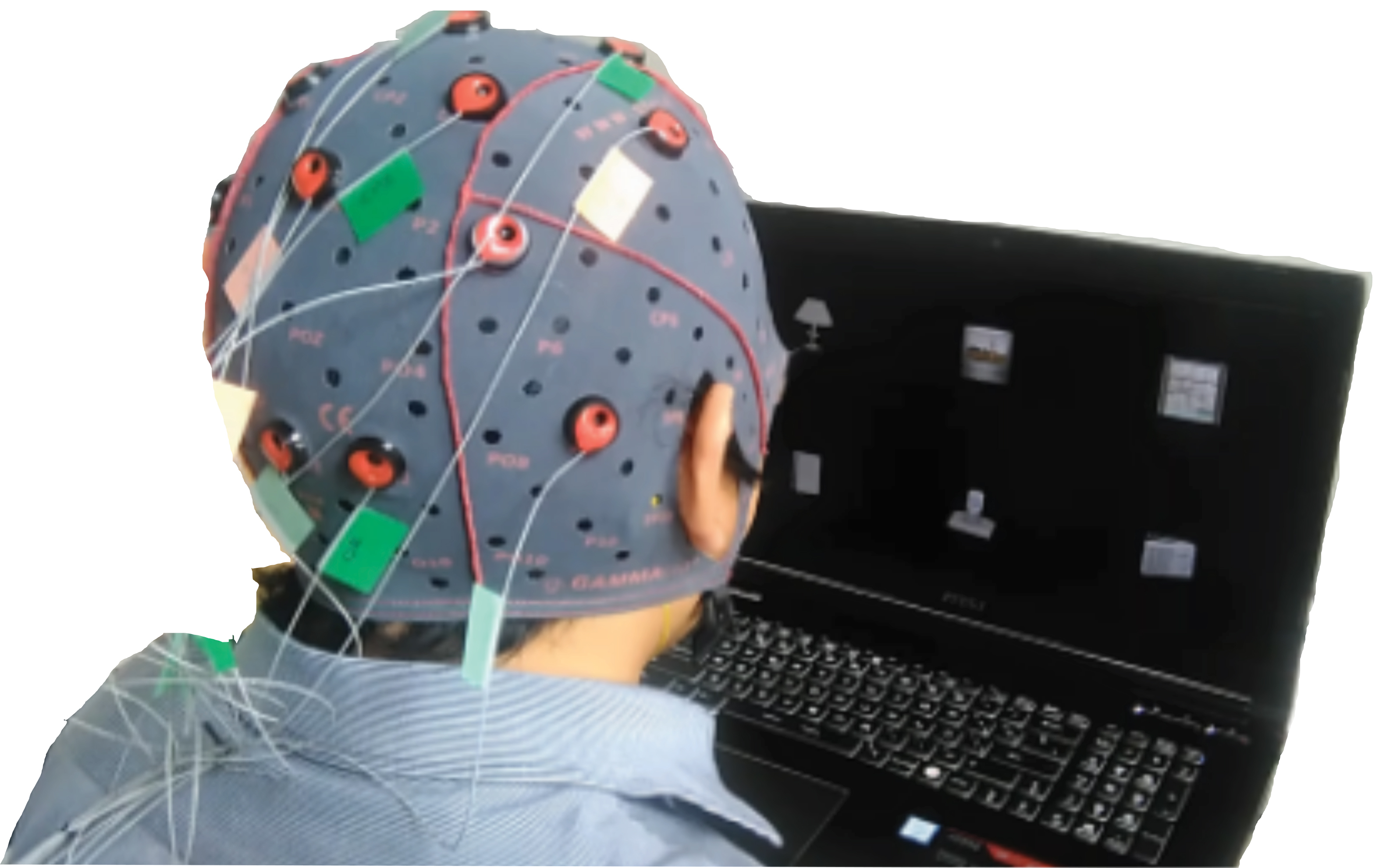}};
   \draw[ultra thick,rounded corners=5pt, blue!50!black, fill = yellow!10]  (9.25,5) -- (8.5,5)  -- (11.45,7.5) -- (14.5,5)-- (13.75,5) -- (13.75,2.75) -- (9.25,2.75) -- cycle;
    \node (P300_GUI) [block, draw=none, fill= none, text width = 6cm, right of = participant, node distance =11.5cm, label=above:Smart Home Interaction]{\includegraphics[scale=0.125, trim = {1cm 1cm 1cm 1cm}, clip= true ] {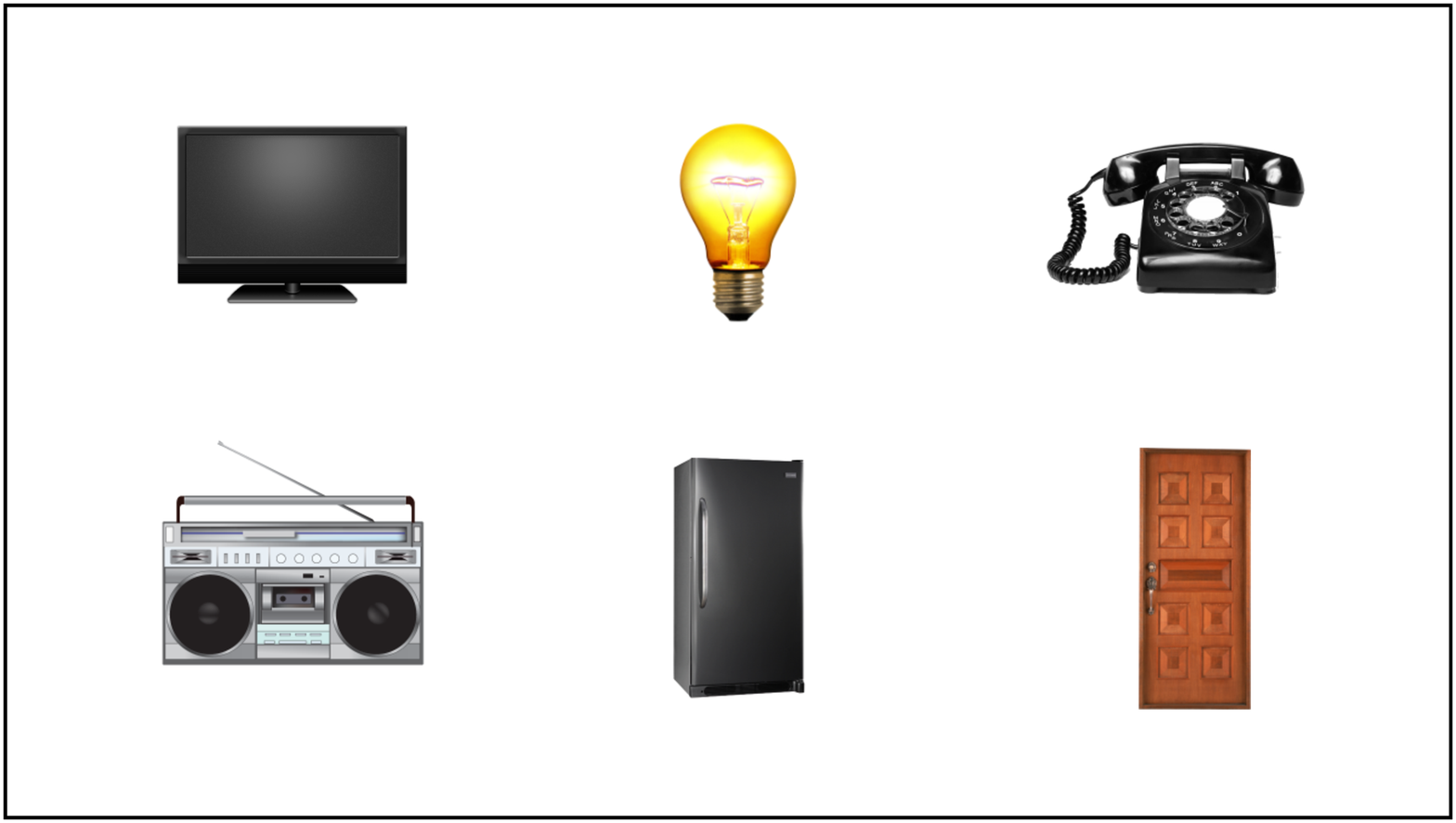}};
    \begin{scope}[x={(P300_GUI.south east)},y={(P300_GUI.north west)}]
        \draw[red,ultra thick,rounded corners] (0.43,0.56) rectangle ++(0.006,0.14);
    \end{scope}
    \node (pre-process) [block, below of = brain_sig, text width = 3cm, fill= green!30, node distance = 4cm] { \begin{tabular}{l}
         Preprocessing 
    \end{tabular} };
    \node (model_FE) [block, fill= red!30, right of = pre-process, text width = 2.5cm, node distance = 4.5cm] {Feature Extraction};
    \node (model_Class) [block, right of = model_FE, text width = 2cm, fill= red!30, node distance = 3.5cm] {Classifier};
    \node (model) [dotted_block, inner ysep=4mm, inner xsep=4mm, fit=(model_FE)(model_Class), label={[yshift=-3cm]Classification Mechanism}] {};
    \node (device) [block,above of = model, text width = 3cm, fill= cyan!20, node distance =  4cm] {\begin{tabular}{c}
         Home \\
         Appliance
    \end{tabular}};
    \node (training) [block, right of = device, text width = 3cm, fill= yellow!30, node distance = 4.5cm] {\begin{tabular}{l}
         Data Analysis \\
         Training Phase
    \end{tabular}};
    \draw[myarrows] (brain_sig)--(pre-process);
    \draw[myarrows] (pre-process)--(model);
    \draw[myarrows] (model)--(device) node [pos=1.1, left, rotate=90, yshift = .7cm] (TextNode1) 
    {\begin{tabular}{l}
    Decoded \\ 
    Intention
    \end{tabular}};
    \draw [myarrows] (model) -|(training);
    \draw [myarrows] ([yshift=0.5cm]participant.west) -| ++ (-2cm,-4.5cm) -- (brain_sig.west);
    \draw [myarrows] ([yshift=0.5cm]participant.east) --([yshift=0.5cm]P300_GUI.west);
    \draw[myarrows, dashed] (device)|-(participant) node [pos=0.45, left, rotate=90, yshift = 0.3cm] (TextNode2) 
    {\begin{tabular}{l}
    Feedback 
    \end{tabular}};
\end{tikzpicture}

%% file: Figures/CMwithFNB.tex
\tikzstyle{block} = [rectangle, draw,  
    text width=2.8cm, text centered, rounded corners, minimum height=1.5cm, fill= none]
\tikzstyle{neuron} = [circle, draw, inner sep = 0cm, font= \tiny,  minimum size=10pt, fill=none]
\tikzstyle{neuron_FNB} = [rectangle, draw,  font= \tiny,  minimum size=10pt, fill=none]
\tikzstyle{dotted_block} = [draw=black!30!white, line width=1pt, dash pattern=on 1pt off 4pt on 6pt off 4pt, inner ysep=1mm,inner xsep=1mm, rectangle, rounded corners ]
\tikzstyle{block_FNB} = [rectangle, draw, text width=.5cm, text badly centered, minimum height = .5 cm, font=\tiny ]
\pgfmathdeclarefunction{gauss}{2}{%
  \pgfmathparse{1/(#2*sqrt(2*pi))*exp(-((x-#1)^2)/(2*#2^2))}%
}
\begin{figure}[!t]
\centering
\begin{tikzpicture}
\node (CM1) [block, fill=red!30, text width =2cm ] {\begin{tabular}{c}
  Base \\ architecture
\end{tabular}};
\node (label_CM1) [block, fill=none, text width =5cm, draw=none, below of = CM1, xshift = .58cm, yshift = -.2cm ] {\begin{tabular}{l}
 (a) Deep learning classifier model without FNB.
\end{tabular}};
\node (CM2) [block, fill=red!30, text width =2cm, below of = CM1, node distance = 4cm ] {\begin{tabular}{c}
  Base \\ architecture
\end{tabular}};
\node (label_CM2) [block, fill=none, text width =5cm, draw=none, minimum height=.8cm, below of = CM2, xshift = .58cm, yshift = -3.8cm, inner sep = 0ex ] {\begin{tabular}{l}
 (b) Deep learning classifier model with FNB.
\end{tabular}};
\node (FC1) [block, fill=blue!20, text width =3.5cm, right of = CM1, node distance = 3.5cm ] {\begin{tabular}{c}
   Fully Connected (FC) \\
   Layers
\end{tabular}};
\node (FC2) [block, fill=blue!20, text width =3.5cm, right of = CM2, node distance = 3.5cm ] {\begin{tabular}{c}
   Fully Connected (FC) \\
   Layers
\end{tabular}};
\node (inp1) [block, draw = none, left of = CM1, node distance=2.5cm, text width =1cm] {Input};
\node (inp2) [block, draw = none, left of = CM2, node distance=2.5cm, text width =1cm] {Input};
\foreach \x in {9}
    \foreach \y in {0,...,4} 
      {\pgfmathtruncatemacro{\label}{\x - 5 *  \y +10}
      \node (f\y) [block_FNB] at (\x,.5*\y-1.5) {};}
\node[block, draw = none, rotate = 90, font = \Huge] at (9, -.7-1.5) {...};
\node(f5)[block_FNB] at (9, -1.3-1.5) {};
\foreach \x in {13}
    \foreach \y in {0,...,4} 
      {\pgfmathtruncatemacro{\label}{\x - 5 *  \y +10}
      \node (g\y) [block_FNB] at (\x,.5*\y-1.5) {};}
\node[block, draw = none, rotate = 90, font = \Huge] at (13, -.7-1.5) {...};
\node(g5)[block_FNB] at (13, -1.3-1.5) {};
\node (MF_block) [block, fill=none, text width =11cm, draw, right of = f5, xshift = 1cm, yshift = 1.8cm, scale = 0.2, minimum height = 5cm]{};
\node (MF) [block, fill=none, text width =11cm, draw=none, right of = f5, xshift = 1.25cm, yshift = 1.5cm, scale = 0.2, minimum height = 5cm] {\begin{axis}[
  no markers, domain=0:13, samples=100,
  axis lines*=left, 
  every axis y label/.style={at=(current axis.above origin),anchor=south},
  every axis x label/.style={at=(current axis.right of origin),anchor=south},
  height=5cm, width=10cm,
  ytick=\empty,
  xtick=\empty,
  enlargelimits=false, clip=false, axis on top
  ]
  \addplot [very thick,cyan!50!black] {gauss(2,1)};
  \addplot [very thick,cyan!50!black] {gauss(4,1)};
  \addplot [very thick,cyan!50!black] {gauss(8,1)};
    \addplot [very thick,cyan!50!black] {gauss(10,1)};
  \addplot[mark=none, dashed, black, ultra thick] coordinates {(5.5,0.18) (6.5,.18)};
\end{axis}};
\node (o0) [neuron, right of = FC1, node distance = 3cm, fill=red!30, yshift = -.5cm] {}; 
\node (o1) [neuron, right of = FC1, yshift = .5cm, node distance = 3cm, fill=green!30] {}; 
\node (c0)[block, draw =none, right of = o0, xshift=0cm, inner sep=1ex, align=flush left, text width = 1.2cm, font=\small\linespread{0.1}\selectfont ] { \textcolor{blue}{Rest of the devices}};
\node (c1)[block, draw =none, right of = o1, xshift=0cm,inner sep=1ex, align=flush left, text width = 1cm, font=\small\linespread{0.1}\selectfont] {\textcolor{blue}{Decoded device}};
\node (FNB_1) [dotted_block, fit= (f5)(f4)(g5)(g4)(MF),inner ysep=4mm, inner xsep=4mm, label= ] {};
\node (label_FNB_1) [block, draw = none, text width = 6cm, below of = FNB_1, yshift=-1.65cm] {(c) Fuzzy Neural Block (FNB).};
\node (FNB) [block, below of = FC2, fill = yellow!30, node distance = 3.5cm, text width = 3cm] {\begin{tabular}{l}
     Fuzzy Neural \\ Block (FNB)
\end{tabular}};
\node (Merge) [block, rounded corners, fill=cyan!20, text badly centered, right of = FC2, text width = 3cm, node distance = 3cm, minimum height = 1 cm, yshift=-1.75cm ]{\begin{tabular}{l}
Merge Layer
\end{tabular}};
\node(On1) [neuron, right of = Merge, fill = green!30, node distance = 3cm, yshift = .5cm]{};
\node(ON)[block, draw=none, right of=On1, node distance = .9cm, xshift=.5cm]{\begin{tabular}{l}
\textcolor{blue}{Decoded device}
\end{tabular}};
\node(On2) [neuron, right of = Merge, fill = red!30, node distance = 3cm, yshift = -.5cm]{};
\node(OFF)[block, draw=none, right of=On2, node distance = .9cm, xshift=.5cm]{\begin{tabular}{l}
\textcolor{blue}{Rest of the devices}
\end{tabular}};
\foreach \h  in {0,...,5} 
     {\draw [-]  (f\h.east)-- (MF_block.west);}
\foreach \h  in {0,...,5} 
     {\draw [-]  (g\h.west)-- (MF_block.east);}
\draw[->, thick] (inp1) -- (CM1);
\draw[->, thick] (inp2) -- (CM2);
\draw[->, thick] (CM1) -- (FC1);
\draw[->, thick] (FC1) -- ++(2.2,0)|-  (o0);
\draw[->, thick] (FC1) -- ++(2.2,0)|-  (o1);
\draw[->, thick] (CM2) -- (FC2);
\draw[->, thick] (CM2) |- (FNB);
\draw[->, thick] (FC2.east) -| (Merge);
\draw[->, thick] (FNB) -| (Merge);
\draw[->, thick] (Merge)-- ++(2.2,0)|- (On1);
\draw[->, thick] (Merge)-- ++(2.2,0)|- (On2);
\end{tikzpicture}
\caption{In the present work, we validate the efficacy of the addition of fuzzy neural block (FNB) by testing the EEG based P300 smart home interaction BCI using classification models with and without the FNB. (a) For 
deep learning classifier model without FNB, the output of the fully connected (FC) layer is fed to the softmax activation layer that outputs the probabilities for each class label, in this case: On and Off. (b) When classification models deep learning classifier model are appended with the FNB, the output of the FC layers is merged with the output of the FNB before input to the softmax activation layer. (c) The FNB flattens the input it receives and outputs values based on eq. (\ref{eq:output_FNB}) using gaussian membership functions.}
\label{fig:CM_withFNB}
\end{figure}

%% file: Figures/Architecture.tex
\tdplotsetmaincoords{60}{40}
\scalebox{.235}{
\hspace{-4cm}
\begin{tikzpicture}[tdplot_main_coords]
\tikzstyle{outer_box} = [thick, fill=lightgray, fill opacity=.15]
\tikzstyle{inner_box} = [thick, fill=yellow!50, fill opacity=.5]
\tikzstyle{edge_fill} = [thick, fill=cerisepink!50, fill opacity=.5]
\node[inner sep=0pt] (EEGsig) at (-12, 0, 20)
    {\includegraphics[scale=1.2, ]{Figures/Figure_14_10_21.eps}};
\draw[->, ultra thick,-triangle 45,
        line width=2mm] (-16, 0, 10.2) |- (-9, 0,5 );
\draw[outer_box] (-12, 0, 0) -- ++(4, 0, 0) -- ++(0, 0, 2) -- ++(-4, 0, 0) -- cycle;
\draw[outer_box] (-12, 0, 0) -- ++(4, 0, 0) -- ++(0, 21, 0) -- ++(-4, 0, 0) -- cycle;
\draw[outer_box] (-12, 0, 0) -- ++(0, 0, 2) -- ++(0, 21, 0) -- ++(0, 0, -2) -- cycle;
\draw[outer_box] (-12, 0, 0) -- ++(0, 0, 2) -- ++(0, 21, 0) -- ++(0, 0, -2) -- cycle;
\draw[outer_box] (-12, 0, 2) -- ++(4, 0, 0) -- ++(0, 21, 0) -- ++(-4, 0, 0) -- cycle;
\draw[outer_box] (-12, 21, 0) -- ++(4, 0, 0) -- ++(0, 0, 2) -- ++(-4, 0, 0) -- cycle;
\draw[inner_box] (-12+2, 0, 0) -- ++(2, 0, 0) -- ++(0, 0, 2) -- ++(-2, 0, 0) -- cycle;
 \draw[inner_box] (-12+2, 0, 0) -- ++(2, 0, 0) -- ++(0, 10.5, 0) -- ++(-2, 0, 0) -- cycle;
\draw[inner_box] (-12+2, 0, 0) -- ++(0, 0, 2) -- ++(0, 10.5, 0) -- ++(0, 0, -2) -- cycle;
\draw[inner_box] (-12+2+2, 0, 0) -- ++(0, 0, 2) -- ++(0, 10.5, 0) -- ++(0, 0, -2) -- cycle;
\draw[inner_box] (-12+2, 0, 0+2) -- ++(2, 0, 0) -- ++(0, 10.5, 0) -- ++(-2, 0, 0) -- cycle;
 \draw[inner_box] (-12+2, 0+10.5, 0) -- ++(2, 0, 0) -- ++(0, 0, 2) -- ++(-2, 0, 0) -- cycle;
\draw[edge_fill] (-12+4, 0, 0) -- ++(8/1.5, 4.25, 1) -- ++(-8/1.5, 6.25, -1)-- cycle ;
\draw[edge_fill] (-12+4, 0, 0) -- ++(0, 0, 2) -- ++ (8/1.5,4.25 ,-1) -- cycle;
\draw[edge_fill] (-12+4, 10.5, 0) -- ++(0, 0, 2) -- ++ (8/1.5,-6.25,-1) -- cycle;
\draw[edge_fill] (-12+4, 0, 2) -- ++(8/1.5, 4.25, -1) -- ++(-8/1.5, 6.25,1) -- cycle;
\draw (-12, 25.5, 0) node[rotate=0,font=\fontsize{32}{38}\sffamily\bfseries]{L1};
\draw (-10, -1.5, 0) node[rotate=-20,font=\fontsize{32}{38}\sffamily\bfseries]{16};
\draw (-13, 0.5, 0) node[rotate=0,font=\fontsize{32}{38}\sffamily\bfseries]{8};
\draw (-7, 2, 0) node[rotate=30,font=\fontsize{32}{38}\sffamily\bfseries]{120};
\draw[outer_box] (2-3, 0, 0) -- ++(1, 0, 0) -- ++(0, 0, 4*1) -- ++(-1, 0, 0) -- cycle;
\draw[outer_box] (2-3, 0, 0) -- ++(1, 0, 0) -- ++(0, 21, 0) -- ++(-1, 0, 0) -- cycle;
\draw[outer_box] (2-3, 0, 0) -- ++(0, 0, 4*1) -- ++(0, 21, 0) -- ++(0, 0, -4*1) -- cycle;
\draw[outer_box] (2+1-3, 0, 0) -- ++(0, 0, 4*1) -- ++(0, 21, 0) -- ++(0, 0, -4*1) -- cycle;
\draw[outer_box] (2-3, 0, 4*1) -- ++(1, 0, 0) -- ++(0, 21, 0) -- ++(-1, 0, 0) -- cycle;
\draw[outer_box] (2-3, 21, 0) -- ++(1, 0, 0) -- ++(0, 0, 4) -- ++(-1, 0, 0) -- cycle;
\draw (-1.5, 23.5, 3) node[rotate=0,font=\fontsize{32}{38}\sffamily\bfseries]{L2};
\draw (-.5, -1.5, 0) node[rotate=-20,font=\fontsize{32}{38}\sffamily\bfseries]{1};
\draw (-2.5, 0.5, 0) node[rotate=0,font=\fontsize{32}{38}\sffamily\bfseries]{16};
\draw (0.75, 2, 0) node[rotate=30,font=\fontsize{32}{38}\sffamily\bfseries]{120};

\draw[inner_box] (2-3, 10.5, 0) -- ++(1, 0, 0) -- ++(0, 0, 4*1) -- ++(-1, 0, 0) -- cycle;
 \draw[inner_box] (2-3, 10.5, 0) -- ++(1, 0, 0) -- ++(0, 10.5/2, 0) -- ++(-1, 0, 0) -- cycle;
\draw[inner_box] (2-3, 10.5, 0) -- ++(0, 0, 4*1) -- ++(0, 10.5/2, 0) -- ++(0, 0, -4*1) -- cycle;
\draw[inner_box] (2+1-3, 10.5, 0) -- ++(0, 0, 4*1) -- ++(0, 10.5/2, 0) -- ++(0, 0, -4*1) -- cycle;
\draw[inner_box] (2-3, 10.5, 0+4*1) -- ++(1, 0, 0) -- ++(0, 10.5/2, 0) -- ++(-1, 0, 0) -- cycle;
 \draw[inner_box] (2-3, 10.5+10.5/2, 0) -- ++(1, 0, 0) -- ++(0, 0, 4*1) -- ++(-1, 0, 0) -- cycle;
\draw[edge_fill] (1+2-3, 10.5, 0) -- ++(8, 4.25/2, 2*1) -- ++(-8, 6.25/2, -2*1)-- cycle ;
\draw[edge_fill] (1+2-3, 10.5, 0) -- ++(0, 0, 4*1) -- ++ (8,4.25/2 ,-2*1) -- cycle;
\draw[edge_fill] (1+2-3, 21-10.5/2, 0) -- ++(0, 0, 4*1) -- ++ (8,-6.25/2,-2*1) -- cycle;
\draw[edge_fill] (1+2-3, 10.5, 4*1) -- ++(8, 4.25/2, -2*1) -- ++(-8, 6.25/2,2*1) -- cycle;
\draw[outer_box] (2+8, 5*2, 0+2) -- ++(1, 0, 0) -- ++(0, 0, 4/2) -- ++(-1, 0, 0) -- cycle;
\draw[outer_box] (2+8, 5*2, 0+2) -- ++(1, 0, 0) -- ++(0, 10/2, 0) -- ++(-1, 0, 0) -- cycle;
\draw[outer_box] (2+8, 5*2, 0+2) -- ++(0, 0, 4/2) -- ++(0, 10/2, 0) -- ++(0, 0, -4/2) -- cycle;
 \draw[outer_box] (2+8+1, 5*2, 0+2) -- ++(0, 0, 4/2) -- ++(0, 10/2, 0) -- ++(0, 0, -4/2) -- cycle;
\draw[outer_box] (2+8, 5*2, 0+4/2+2) -- ++(1, 0, 0) -- ++(0, 10/2, 0) -- ++(-1, 0, 0) -- cycle;
 \draw[outer_box] (2+8, 5*2+10/2, 0+2) -- ++(1, 0, 0) -- ++(0, 0, 4/2) -- ++(-1, 0, 0) -- cycle;
 
\draw (12, 13, 5+1.5) node[rotate=0,font=\fontsize{32}{38}\sffamily\bfseries]{L3};
\draw (8+1+1+1, 9, 0+2) node[rotate=-20,font=\fontsize{32}{38}\sffamily\bfseries]{1};
\draw (4+8, 10, 0+2) node[rotate=30,font=\fontsize{32}{38}\sffamily\bfseries]{30};
\draw (8.5, 11, 0+2) node[rotate=0,font=\fontsize{32}{38}\sffamily\bfseries]{8};

\draw[inner_box] (2+8, 5*2+5/2, 0+2) -- ++(1, 0, 0) -- ++(0, 5/4, 0) -- ++(-1, 0, 0) -- cycle;
\draw[inner_box] (2+8, 5*2+5/2, 0+2) -- ++(0, 0, 4/2*1) -- ++(0, 5/4, 0) -- ++(0, 0, -4/2*1) -- cycle;
\draw[inner_box] (2+8, 5*2+5/2, 0+2) -- ++(1, 0, 0) -- ++(0, 0, 4/2*1) -- ++(-1, 0, 0) -- cycle;
\draw[inner_box] (2+8, 5*2+5/2+5/4, 0+2) -- ++(1, 0, 0) -- ++(0, 0, 4/2*1) -- ++(-1, 0, 0) -- cycle;
\draw[inner_box] (2+8+1, 5*2+5/2, 0+2) -- ++(0, 0, 4/2*1) -- ++(0, 5/4, 0) -- ++(0, 0, -4/2*1) -- cycle;
\draw[inner_box] (2+8, 5*2+5/2, 0+4/2*1+2) -- ++(1, 0, 0) -- ++(0, 5/4, 0) -- ++(-1, 0, 0) -- cycle;

\draw[edge_fill] (2+8+1, 5*2+5/2, 0+2) -- ++(8/2, 5/8, 2/2*1) -- ++ (-8/2, 5/8,-2/2*1) -- cycle;
\draw[edge_fill] (2+8+1, 5*2+5/2, 4/2*1+2) -- ++(8/2, 5/8, -2/2*1) -- ++ (-8/2, -5/8,-2/2*1) -- cycle;
\draw[edge_fill] (2+8+1, 5*2+5/2+5/4, 4/2*1+2) -- ++(8/2, -5/8, -2/2*1) -- ++ (-8/2, 5/8,-2/2*1) -- cycle;
\draw[edge_fill] (2+8+1, 5*2+5/2, 4/2*1+2) -- ++(8/2, 5/8, -2/2*1) -- ++(-8/2, 5/8, 2/2*1) -- cycle;
\draw[ultra thick, - ] (16, 12, 7.5) -- ++(0, 0, -3.5);
\draw[ultra thick, ->, -triangle 45,
        line width=1mm] (16, 12,7.5)  -- ++(5.5, 1, 1.75);
\end{tikzpicture}}

%% file: Figures/CNN_classification.tex
\begin{tikzpicture}[>=latex, every label/.append style={text=black, font=\tiny}]
\hspace{-0.35cm}
\tikzstyle{block} = [rectangle, draw, text width=.5cm, text badly centered, minimum height = .5 cm, font=\tiny ]
\tikzstyle{neuron} = [circle, draw, inner sep = 0cm, font= \tiny,  minimum size=10pt]
\tikzstyle{dotted_block} = [draw=black!30!white, line width=1pt, dash pattern=on 1pt off 4pt on 6pt off 4pt,
            inner ysep=1mm,inner xsep=4mm, rectangle, rounded corners ]
\foreach \x in {0}
    \foreach \y in {0,...,5} 
      {\pgfmathtruncatemacro{\label}{\x - 5 *  \y +10}
      \node (f\x\y) [block] at (\x,.5*\y) {};}
\node[block, draw = none, rotate = 90, font = \Huge] at (0, -.7) {...};
\node(f06)[block] at (0, -1.3) {};
\node(L4)[block, draw=none, above of=f04]{L4};
\node (TCN) [block, rounded corners, fill=brown!50, right of = f02, text width = 1cm, node distance = 1.5cm, minimum height = 1 cm, label = L5]{\begin{tabular}{l}
     F = 6  \\
     K = 2 
\end{tabular}};
\node(node_TCN)[block, draw=none, below of=TCN, node distance = .8cm]{TCN};
\node (LSTM1) [block, rounded corners, fill=orange!50, right of = TCN, text width = 1cm, node distance = 1.5cm, minimum height = 1 cm,label = L6]{\begin{tabular}{l}
    N = 30 
\end{tabular}};
\node(node_LSTM1)[block, draw=none, below of=LSTM1, node distance = .8cm,xshift=-.2cm]{LSTM1};
\node (LSTM2) [block, rounded corners, fill=orange!50, right of = LSTM1, text width = 1cm, node distance = 1.5cm, minimum height = 1 cm]{\begin{tabular}{l}
    N = 30 
\end{tabular}};
\node(node_LSTM2)[block, draw=none, below of=LSTM2, node distance = .8cm,xshift=-.2cm]{LSTM2};
\node(node2_LSTM2)[block, draw=none, above of=LSTM2, node distance = .7cm, xshift=-.4cm]{\begin{tabular}{l}
L7
\end{tabular}};
\node (FC) [block, rounded corners, fill=blue!20,  left of = LSTM2, text width = 1.4cm, node distance = 1.5cm, minimum height = 1 cm,label = {[xshift=-.5cm, yshift=0cm]L8}, yshift = 1.8cm]{\begin{tabular}{l}
Fully\\Connected
\end{tabular}};
\node (Fuzzy) [block, rounded corners, fill=yellow!30, text badly centered, right of = LSTM2, text width = 1.7cm, node distance = 1.5cm, minimum height = 1 cm,label = {[xshift=.5cm, yshift=0cm]L8},  yshift = 1.8cm]{\begin{tabular}{l}
Fuzzy Neural \\ Block (FNB)
\end{tabular}};
\node (Merge) [block, rounded corners, fill=cyan!20, text badly centered, above of = LSTM2, text width = 1.2cm, node distance = 3.5cm, minimum height = 1 cm ]{\begin{tabular}{l}
\parbox{.15cm}Merge\\Layer
\end{tabular}};
\node(node_Merge)[block, draw=none, above of=Merge, node distance = .7cm, xshift=0cm]{\begin{tabular}{l}
L9
\end{tabular}};
\node(n1) [neuron, right of = Merge, fill = green!20, node distance = 1.5cm, yshift = .5cm, label = L10]{};
\node(ON)[block, draw=none, right of=n1, node distance = .9cm, xshift=-.5cm]{\includegraphics[scale=0.1,trim={15cm 10cm 15cm 1cm},clip]{Figures/p300_screen_stimulus.eps}};
\node(n2) [neuron, right of = Merge, fill = red!20, node distance = 1.5cm, yshift = -.5cm]{};
\node(OFF)[block, draw=none, right of=n2, node distance = .9cm, xshift=-.5cm]{\includegraphics[scale=0.1,trim={15cm 11cm 15cm 2cm},clip]{Figures/p300_screen_stimulus.eps}};
\node(OFF_northeast) [block, draw=none, above of = OFF, xshift = .35cm, node distance = 0.5cm] {};
\node(OFF_southwest) [block, draw=none, below of = OFF, xshift = -.35cm, node distance = 0.5cm] {};
\draw[-] (f00.east)--(TCN.west);
\draw[-] (f01.east)--(TCN.west);
\draw[-] (f02.east)--(TCN.west);
\draw[-] (f03.east)--(TCN.west);
\draw[-] (f04.east)--(TCN.west);
\draw[-] (f05.east)--(TCN.west);
\draw[-] (f06.east)--(TCN.west);
\draw[->] (TCN)--(LSTM1);
\draw[->] (LSTM1)--(LSTM2);
\draw[->] (LSTM2.north)-- ++(0,0.4)-|(Fuzzy.south);
\draw[->] (LSTM2.north)-- ++(0,0.4)-|(FC.south);
\draw[->] (FC.north)--++(0,0.3)-|(Merge);
\draw[->] (Fuzzy.north)--++(0,0.3)-|(Merge);
\draw[->] (Merge.east)--++(.3,0)|- (n1);
\draw[->] (Merge.east)--++(.3,0)|- (n2);
\draw[-, red!60!brown, rotate=45 ,ultra thick] (OFF_northeast) --(OFF_southwest);
\end{tikzpicture}